\pdfoutput=1

\documentclass[11pt]{article}

\usepackage[preprint]{acl}

\usepackage{times}
\usepackage{graphicx}
\usepackage{array}
\usepackage{latexsym}
\usepackage{float}
\usepackage{hyperref}
\usepackage{subfigure}
\usepackage[table]{xcolor}  

\usepackage{enumitem} 

\usepackage[most]{tcolorbox}  
\usepackage{xcolor}           

\usepackage{booktabs}
\usepackage{siunitx} 

\definecolor{lightblue}{RGB}{143,217,251} 
\definecolor{lightpink}{RGB}{211,211,255} 
\definecolor{lightgreen}{RGB}{200,255,200}  
\definecolor{lightyellow}{RGB}{252, 212, 243}

\usepackage[T1]{fontenc}

\usepackage[utf8]{inputenc}

\usepackage{multirow}

\usepackage{microtype}

\usepackage{inconsolata}

\usepackage{tabularx}

\usepackage{graphicx}
\usepackage{xspace}
\usepackage{multirow}

\newcommand{\approach}{\textit{GRAVITY}\xspace}
\newcommand{\values}{\colorbox{lightpink}{Values and Beliefs}}
\newcommand{\personality}{\colorbox{lightblue}{Personality Traits}}
\newcommand{\interests}{\colorbox{lightyellow}{Interests}}

\usepackage{xargs} 
\usepackage{hyperref}
\usepackage[colorinlistoftodos,prependcaption,textsize=tiny]{todonotes} 
\newcommandx{\jz}[2][1=]{\todo[linecolor=blue,backgroundcolor=blue!25,bordercolor=blue,#1]{Jieyu: #2}}

\title{GRAVITY: A Framework for Personalized Text Generation via Profile-Grounded Synthetic Preferences}

\author{
Priyanka Dey\textsuperscript{$\spadesuit$} \quad
Daniele Rosa\textsuperscript{$\clubsuit$} \quad
Wenqing Zheng\textsuperscript{$\clubsuit$} \quad
Daniel Barcklow\textsuperscript{$\clubsuit$} \quad \\
\textbf{Jieyu Zhao}\textsuperscript{$\spadesuit$} \quad
\textbf{Emilio Ferrara}\textsuperscript{$\spadesuit$} \\ 
\textsuperscript{$\spadesuit$}University of Southern California \quad \textsuperscript{$\clubsuit$}Capital One Research \\
\texttt{\{deyp\}@usc.edu}
}

\begin{document}
\maketitle
\begin{abstract}

Personalization in LLMs often relies on costly human feedback or interaction logs, limiting scalability and neglecting deeper user attributes. To reduce the reliance on human annotations, we introduce \approach (\textbf{G}enerative \textbf{R}esponse with \textbf{A}ligned \textbf{V}alues, \textbf{I}nterests, and \textbf{T}raits of \textbf{Y}ou), a framework for generating \textbf{synthetic, profile-grounded preference data} that captures users’ interests, values, beliefs, and personality traits. By integrating demographic, cultural, and psychological frameworks—including Hofstede’s cultural dimensions, Schwartz’s basic values, the World Values Survey, and Big Five OCEAN traits, \approach synthesizes preference pairs to guide personalized content generation. We evaluate \approach on book descriptions for 400 Amazon users, comparing it to prompt-based conditioning, standard fine-tuning, and naive synthetic pair generation. Profile-grounded synthetic data consistently improves generation, especially across multiple cultures (USA, Brazil, Japan, India), achieving over 4\% higher preference gains across baselines, with user studies showing that \approach outputs are preferred over 86\% of the time. Our results show that scenario-grounded synthetic data can capture richer user variation, reduce reliance on costly annotation, and produce more engaging, user-centered content, offering a scalable path for LLM personalization.\footnote{Code is available: \href{https://github.com/limenlp/GRAVITY}{https://github.com/limenlp/GRAVITY}.}
\end{abstract}

\section{Introduction}
\label{sec:intro}

\begin{figure*}[!t] \centering \includegraphics[width=0.8\textwidth]{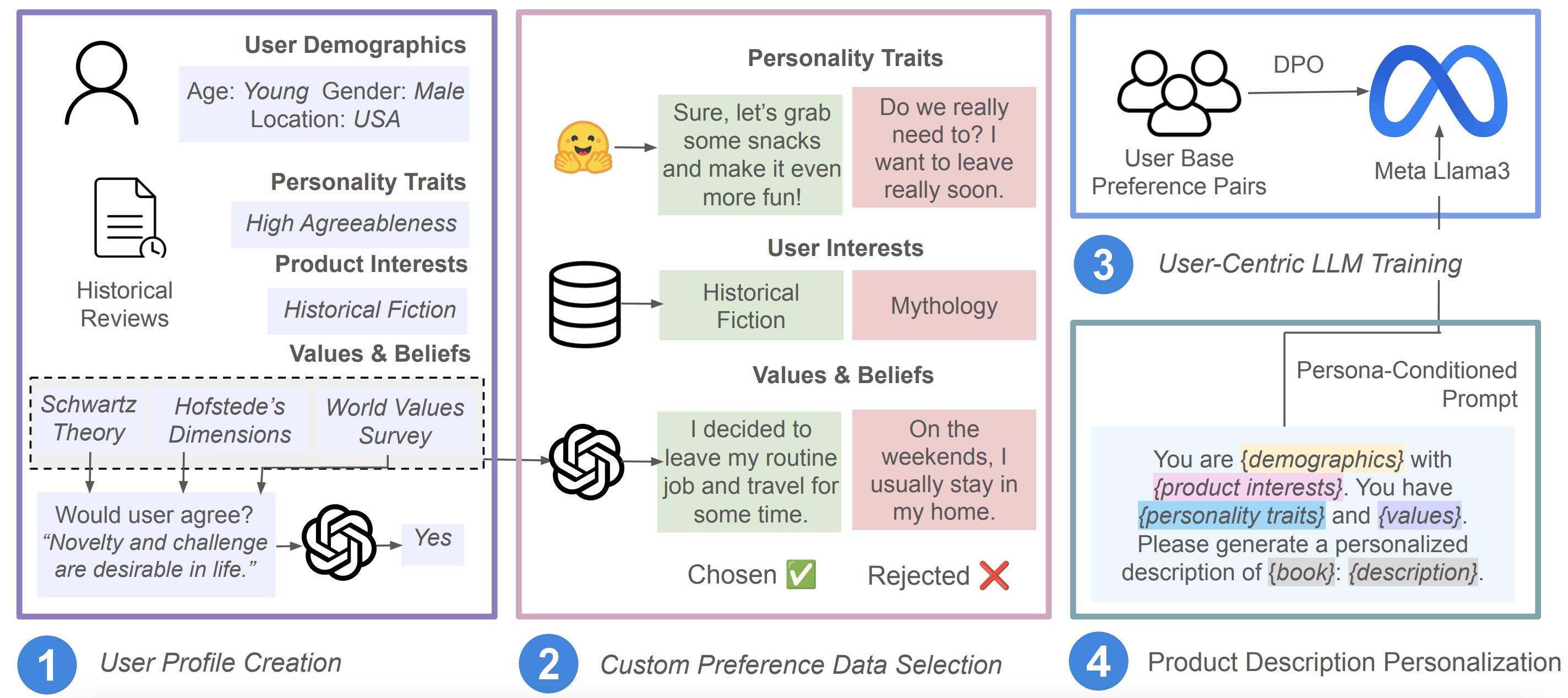} \caption{Our approach \approach consists of four stages: (1) \textit{\textbf{User Profile Creation}}, (2) \textit{\textbf{Custom Preference Data Selection}}, (3) \textit{\textbf{User-Centric LLM Training}}, and (4) \textit{\textbf{Product Description Personalization}}. In stage 1, we extract information about the user including \textit{explicit} values such as demographic attributes: age, location, and gender and users' interests and \textit{implicit} values including their personality traits and their values and beliefs (extracted based on seed statements generated from various psychological and cultural frameworks). In stage 2, During stage 1, we generate a candidate pool of scenarios for values and beliefs based on Stage 1 using GPT-4o. We then generate a custom set of chosen/reject preference pairs for users spanning three facets: user interests, personality, and values using a combination of the Amazon Reviews dataset, personality SJTs (\textit{TRAIT} and \textit{Big5Chat}), and the candidate pool of generated scenarios. In stage 3, we preference tune \textit{Llama} with the users' preference pairs, and finally, in Stage 4, we generate the personalized description by prompting this tuned model with user profile attributes.} \label{fig:approach} \end{figure*}

Personalization has become a critical frontier for LLMs \cite{jang2023personalized, chen2024persona, zhang2024personalization, }. While recent advances enable fluent and contextually relevant text generation, outputs often remain generic, overlooking individual differences in taste, style, and preference \cite{zhang-etal-2025-personalized, moorjani-etal-2022-audience}. In domains such as book recommendations, this gap is particularly visible—two users may value entirely different aspects of the same content, with one drawn to narrative structure and cultural context and the other to character development and personal resonance. Generic descriptions risk alienating users by failing to capture what truly matters \cite{yunusov2024mirrorstoriesreflectingdiversitypersonalized, cai-etal-2023-generating}.

Prior work in personalization often relies on human-annotated preference data, whether through reinforcement learning from human feedback (RLHF) \cite{kirk2024prism, poddar2024personalizing}, preference modeling \cite{lee2024aligning, zhong2024panacea, zheng2025tuningfreellmbuildstrong}, or profile-conditioned prompting \cite{zhang2024p4, lyu2023llm}. While effective, large-scale annotation is costly and difficult to scale, limiting the breadth of user attributes that can be incorporated. Moreover, existing approaches frequently reduce personalization to narrow signals such as demographics or explicit traits, overlooking deeper dimensions of user engagement.

To combat this challenge, we introduce \approach (\textbf{G}enerative \textbf{R}esponse with \textbf{A}ligned \textbf{V}alues, \textbf{I}nterests, and \textbf{T}raits of \textbf{Y}ou), a framework for creating \textbf{synthetic, profile-grounded preference data} that can be used to fine-tune LLMs for more effective personalization. As a case study, we focus on personalizing book descriptions using users from the Amazon Book Reviews dataset \cite{hou2024bridging}. Instead of collecting explicit preference annotations, we construct synthetic preference pairs grounded in well-established psychological and cultural frameworks—including Hofstede’s cultural dimensions \cite{hofstede1983national}, Schwartz’s theory of basic values \cite{schwartz2012overview}, the World Values Survey \cite{Haerpfer2024-jz}, and the Big Five OCEAN traits \cite{goldberg2013alternative}. This allows us to capture variation in user interests, values, beliefs, and personality traits, providing a richer basis for personalization than demographics alone. We then fine-tune \texttt{Llama-3.1-8B-Instruct} with Direct Preference Optimization (DPO) \cite{rafailov2023direct}, aligning generations with these profile-derived preferences.

We evaluate \approach against several baselines, including prompt-based conditioning, standard supervised fine-tuning, and a naive DPO approach using synthetic pairs without structured profiles. Our results show that profile-grounded synthetic data consistently improves generation, achieving over 4\% higher preference scores across baselines. User studies further show that \approach generations are preferred over 86\% of the time.

While our results do not imply that synthetic data can replace human feedback, they suggest that carefully designed, structured, scenario-based synthetic pipelines can reduce annotation needs while capturing various user attributes, offering a scalable path for aligning LLM-generated content with what users actually find engaging. Our contributions are summarized:

\begin{itemize}
    \item \textbf{\approach:} We introduce \approach, a multi-step framework for personalized content generation using \textbf{synthetic, profile-grounded user data}, capturing values, interests, and personality traits while reducing reliance on costly human annotation.
    \item \textbf{Profile Modeling:} We develop a pipeline integrating psychological and cultural frameworks (Hofstede, Schwartz, WVS, OCEAN) to generate customized preference data that reflects diverse user profiles.
    \item \textbf{Comprehensive Evaluation:} We assess our method using both automatic metrics and a user study, measuring preference alignment and user-perceived engagingness of the generated content.
    \item \textbf{Data Efficiency:} We show that profile-grounded synthetic preference data yields measurable improvements over prompt-based, SFT, and naive DPO baselines.
\end{itemize}

\section{Background}
\label{sec:background}

Personalization is central to making generated content engaging and relevant: attributes such as user interests, style preferences (especially through personality) \cite{nguyen2018user, dey2025can}, or cultural background \cite{matz2024potential, joshi2025improving, liu2025can} can strongly shape whether stories and characters resonate with readers \cite{wozniak2024personalized, yang2023palr}. Recent work leverages LLMs for personalization across both direct text generation and downstream applications like recommendation \cite{jiang2023personallm, jang2023personalized}. Broadly, methods fall into two streams: (a) personalized generation, where models adapt outputs via prompting, user profiles, or fine-tuning \cite{peng2024review, li2024personalized}; and (b) LLMs for personalized tasks, such as recommendation, reasoning over user histories, or tailoring explanations \cite{bismay2024reasoningrec, shao2024ulmrec}. Hybrid approaches increasingly combine these, generating user-specific content that simultaneously improves alignment and task performance \cite{shenfeld2025language, zhong2024panacea}.

Within generation, personalization extends beyond static profile conditioning to new task formulations. Work on summarization, for example, uses multi-agent pipelines to refine drafts according to user preferences \cite{xiao2023personalized}, while headline generation has incorporated user-history modeling \cite{song2023general}. Similarly, recommender systems use LLMs to expand sparse item text or construct structured user/item profiles from histories, enabling personalized reviews and explanations \cite{lyu2023llm, acharya2023llm}.

Finally, advances in alignment and evaluation point toward more fine-grained personalization. Extensions of RLHF introduce per-user reward components or factorized preference functions \cite{li2024personalized, shenfeld2025language, poddar2024personalizing}, while controllable preference vectors allow multi-objective adaptation \cite{zhong2024panacea}. Benchmarks like LaMP \cite{salemi2023lamp} and newer evaluations on role-playing, user modeling, and cultural adaptation \cite{tseng2024two, zhang2024personalization} reflect this growing focus on personalization as a key axis of LLM evaluation.

\section{\approach}
\label{sec:task}

In this section, we present our framework \approach for generating personalized content. We first describe product description personalization (\S\ref{sec:prelims}), then our user profile pipeline (\S\ref{sec:user_profiles}) and synthetic preference data generation (\S\ref{sec:custom_preferences}). Finally, we outline user-centric LLM training (\S\ref{sec:training}) and personalized description generation (\S\ref{sec:generation}). For evaluation, we focus on generating personalized book descriptions for 400 Amazon readers from diverse cultural backgrounds; additional details on the users and dataset are provided in \S\ref{app:dataset}.

\subsection{Task Formulation}
\label{sec:prelims}


We formulate personalization of product descriptions as a conditional generation problem grounded in structured user profiles. Let $\mathcal{U} = \{u_1, u_2, \dots, u_{|\mathcal{U}|}\}$ denote the set of users, and $\mathcal{B} = \{b_1, b_2, \dots, b_{|\mathcal{B}|}\}$ denote the set of products. Each user $u \in \mathcal{U}$ has a historical review sequence 
\[
R_u = (r_{u1}, r_{u2}, \dots, r_{u|R_u|}),
\] 
where $r_{ui}$ is the textual review for product $b_i \in \mathcal{B}$. 

From these reviews and basic demographic information, we curate a structured user profile $P_u$, which includes four components:  
(1) Demographics (e.g., age, gender, location),  
(2) Product interests,  
(3) Intrinsic values and core beliefs, and  
(4) Personality traits.  

Each product $b_n \in \mathcal{B}$ is associated with a generic, non-personalized description $d_n$. Given a target user $u$ and product $b_n$, the goal is to generate a personalized description $\hat{B}_n$ that better aligns with the user’s profile $P_u$. Formally, this can be written:
\[
\hat{B}_n \sim P(\mathcal{D} \mid P_u, d_n; \theta),
\]
where $\mathcal{D}$ denotes the space of possible product descriptions and $\theta$ represents the model parameters. The personalized generation model can be represented as:
\[
\hat{B}_n = F_\theta(P_u, d_n),
\]
where $F_\theta$ is the conditional generation model that incorporates both structured user information and the generic product description. The output $\hat{B}_n$ is a personalized description that reflects the user’s preferences and traits while preserving the content of the original product description.  

\begin{table*}[t]
\centering
\scriptsize
\setlength{\tabcolsep}{6pt} 
\begin{tabular}{>{\raggedright\arraybackslash}p{4cm} >{\raggedright\arraybackslash}p{9cm} > {\raggedright\arraybackslash}p{2cm}}
\toprule
\textbf{Source} & 
\textbf{Seed Statement} & 
\textbf{Total Count}
\\
\addlinespace
\midrule
\textbf{Hofstede's Cultural Dimensions} \newline \cite{hofstede1983national} & Planning for the future and working toward long-term goals is more important than immediate rewards. & 10 \\
\midrule
\textbf{Schwartz' Theory of Basic Values} \newline \cite{schwartz2012overview} & It is important to me to make my own choices and decisions and think independently, even if others may disagree. & 10 \\ \midrule
\textbf{World Values Survey} \newline \cite{Haerpfer2024-jz} & I believe hard work doesn’t generally
bring success—it’s more a matter
of luck and connections. & 130 \\
\bottomrule
\end{tabular}
\vspace{-1mm}
\caption{Example seed statements generated from various cultural and psychological constructs.}
\label{tab:example_seeds}
\end{table*}

\subsection{Creating User Profiles}
\label{sec:user_profiles}

Since personalization relies on a nuanced understanding of the user, the first step in \approach is to extract salient user information and construct a structured profile. As discussed in \S\ref{sec:background}, attributes such as demographics, interests, values, and personality traits can strongly influence how engaged and connected a user feels with generated content.

Each user profile in our framework consists of four key components: (1) Demographics (gender, age, location), (2) \interests, (3) \values, (4) \personality. Demographic information is obtained directly from user-provided data when available. When gender or age is missing, we follow prior work that has shown strong performance in demographic estimation and apply trained DeBERTa models \cite{kaantureyyen2023deberta_age, kaantureyyen2023deberta_gender}. To reduce noise and sparsity, we bin ages into three groups: young (<30), middle-aged (31–60), and senior (>60).



\interests capture a user's content preferences, i.e., all product categories the user might be interested in or frequently engage with. the book topics and categories they most frequently engage with. These interests can be extracted based on users' product purchase and review history, combined with their explicitly selected preferences. Specifically, for our case study on Amazon readers, we extract users' most frequent book genres that account for 10\% or more of their reviews\footnote{This threshold ensures that the selected genres reflect substantial and consistent engagement rather than incidental activity.}.

 \values capture a user's implied norms and ideals across domains such as culture, religion, ethics, politics, and society. To extract these, we adopt a multi-step LLM-based approach (see Figure \ref{fig:approach}). We first construct 150 \textit{seed value statements}, drawing from established cultural theories (Hofstede's cultural dimensions, Schwartz’s theory of basic values) and large-scale value surveys (World Values Survey). Table \ref{tab:example_seeds} shows representative examples. We then prompt an LLM (GPT-4o) with each seed statement and the user’s full set of product reviews, asking it to infer whether the user likely \textit{supports}, \textit{opposes}, or is \textit{neutral} toward the statement. This yields a structured profile of the user’s values and beliefs.

\personality capture a user’s underlying preferences and tendencies, shaping both their writing style and the kinds of story elements they are likely to connect most with. To infer these traits, we use \texttt{PersonalityLM} \cite{wang2024continuous}, a fine-tuned RoBERTa classifier to predict high/low levels for each of the Big Five (OCEAN) traits. In our study, we estimate user personalities using each user's historical book reviews. 

\begin{table*}
\centering
\scriptsize
\begin{tabular}{
    >{\centering\arraybackslash}m{4.5cm}  
    >{\centering\arraybackslash}m{2.25cm}  
    >{\centering\arraybackslash}m{2.25cm}  
    >{\centering\arraybackslash}m{3cm}  
    >{\centering\arraybackslash}m{1.75cm}  
}
\toprule
\textbf{Personalization Method} & \textbf{Top-1 WinRate (\%)} & \textbf{Preference Gain (\%)} & \textbf{Interestingness Score} & \textbf{Text Similarity} \\
\midrule
\textit{Original} & 0.75 & - & 3.74 & - \\
\textit{BaseRewrite} & 1.25 & 63.5 & 3.81 & 72.8 \\
\textit{DemoBased} & 3.25 & 68.5 & 3.69 & 72.4 \\
\textit{UserSummary} & 9.25 & 72.0 & 3.98 & 71.0 \\
\textit{LaMP} & 10.0 & 75.5 & \colorbox{lightgreen}{4.06} & 72.8 \\
\textit{TriAgent} & 7.0 & 73.25 & 3.95 & 70.37 \\
\textit{UserSFT} & 19.5 & 74.25 & 3.91 & \colorbox{lightgreen}{79.92} \\
\textit{PrefAlign} & 22.25 & 79.0 & 4.02 & 79.48 \\
\approach (Ours) & \colorbox{lightgreen}{26.75} & \colorbox{lightgreen}{82.5} & 4.03 & 73.2 \\
\bottomrule
\end{tabular}
\vspace{-1mm}
\caption{We compare several baselines, including existing personalization methods using prompt-based techniques and multi-agents, to our method \approach. We report aggregated top-1 win rates, preference gains, interestingness scores, and text similarity scores over all selected users. Win rates, preference gains, and interestingness scores are measured with an LLM-judge (GPT-4o). We find that our method \approach achieves the highest personalization metrics (↑ WinRate and Preference Gain and comparable Interestingness scores) and comparable text similarity scores to other baseline approaches. The original book description has a win rate below 1\%, showing personalization methods generate more engaging descriptions.}
\label{tab:main_results}
\end{table*}

\subsection{Generating Custom Preferences}
\label{sec:custom_preferences}

Based on users’ generated profiles, the next step in \approach is to construct custom preference data that guides model adaptation toward personalized product descriptions. Rather than relying on generic preference statements, we generate scenario-based data that captures how a user’s interests, values and beliefs, and personality traits shape their judgments in concrete contexts. While the underlying pool of scenarios remains consistent across users, each profile produces a unique mapping of \textit{chosen} and \textit{rejected} labels, ensuring that supervision reflects their individual preferences. This scenario-driven approach provides richer, more discriminative signals than abstract descriptions, enabling the model to learn finer-grained distinctions in how different users engage with content \cite{singh2025fspofewshotpreferenceoptimization, huang2023learning}.

\interests We construct two types of interest-based preference data: (1) \textit{Category}, which contrasts broad product categories (e.g., romance vs. biographies), and (2) \textit{Summary}, which contrasts detailed book descriptions. For our study of personalized book descriptions, we begin with 382 categories from the Amazon Reviews dataset and select 3 representative, highly rated books per category. For each user, we identify their top 3–5 genres (\S\ref{sec:user_profiles}) and, using SentenceBERT embeddings with cosine distance, select the 3 most distinct categories. We then generate \textit{Category} preference pairs (chosen vs. rejected) for the user's preferred genres. For \textit{Summary} data, we form preference pairs from book descriptions of the representative titles in the user’s top and most distinct genres.

\values To generate preference data about user's values and beliefs, we first synthetically generate scenarios using GPT-4o for each seed value statement (\S\ref{sec:user_profiles}). For each seed, GPT-4o produces 3 unique pairs of scenarios, where the first scenario strongly aligns with the statement and the second contradicts it, yielding 450 scenario pairs in total. To generate user-specific preference labels, we utilize each user profile's extracted set of values. For a given seed statement, if the user supports it, all generated aligned scenarios are marked \textit{chosen}; if the user opposes it, aligned scenarios are marked \textit{rejected}\footnote{Scenarios from \textit{neutral} seed statements are omitted from the user's preference dataset.}.

\personality We leverage two situation-based personality questionnaires, TRAIT (scenarios) \cite{Lee2024TRait} and Big5Chat (dialogues) \cite{li2024big5chat}, each containing questions with binary-level answers for the \textit{OCEAN} traits. For each user, we randomly select 150 question–answer pairs from each dataset. For each pair, the answer consistent with the user’s OCEAN trait profile is labeled \textit{chosen}, and the alternative labeled \textit{rejected}.

\subsection{User-Centric LLM Training}
\label{sec:training}

For our task of personalized book descriptions, we generate approximately 1,000 customized preference pairs across all three dimensions for each user (400K for user base). Using this data, we utilize Direct Preference Optimization (DPO) \cite{rafailov2023direct} to generate a single personalized LLM which learns the behaviors and styles of the users in our dataset. Each training instance conditions the model on a user’s demographic, personality, values, and interests, enabling generalization across diverse profiles. At inference, we provide the target user’s profile to generate personalized book descriptions. For evaluation, we sample a highly rated title from the user’s preferred genres, excluding books they have already reviewed. Additional training details are in \S\ref{sec:exp_setup}.

\subsection{Product Description Personalization}
\label{sec:generation}

To generate personalized descriptions, we follow prior work \cite{santurkar2023whose, hwang2023aligning} by conditioning model responses on the user’s profile. Each profile, as described in \S\ref{sec:user_profiles}, includes demographics, genre interests, personality traits, and personal values. Full prompt templates for description generation are provided in \S\ref{app:personalization_prompt}. For our task, we select a target book description to personalize from a randomly chosen highly rated title within the user’s top genres, ensuring the user has not previously purchased or reviewed the book.
\section{Experiments}
In this section, we outline our experimental design for personalized text generation (\S\ref{sec:exp_setup}), including the training pipeline for user-centric LLMs, the baselines used for comparison, and evaluation strategies. We then present results across all methods (\S\ref{sec:results}) and conduct ablations to isolate the contribution of each preference dataset to personalization performance (\S\ref{sec:ablations}). Finally, we compare two training strategies for incorporating user preferences—standard supervised fine-tuning (SFT) and Direct Preference Optimization (DPO)—to assess their relative effectiveness (\S\ref{sec:sft_results}).

\begin{table*}[t]
\centering
\scriptsize
\setlength{\tabcolsep}{6pt} 
\begin{tabular}{>{\raggedright\arraybackslash}p{2.5cm} >{\raggedright\arraybackslash}p{12.75cm}}
\toprule
\textbf{Persona} & 
\textit{Country}: India \newline
\textit{Age}: Young \newline
\textit{Gender}: Female \newline
\interests: Young Adult, Romance, Fiction \newline
\values: Friends are an important part of life.; Excitement, novelty, and change are desirable in life. \newline
\personality: High Openness, High Extroversion \\
\addlinespace
\midrule
\textbf{Original Description} & \textit{Our Chemical Hearts} follows Henry Page, a teenager who’s never fallen in \colorbox{lightyellow}{love}—until Grace Town joins his class. Paired to work on the school newspaper, he discovers the excitement and challenges of \colorbox{lightyellow}{first love}, and the ways it can change a person. \\
\midrule
\textbf{LaMP} & \textit{Our Chemical Hearts} follows Henry Page, a \colorbox{lightyellow}{teenager} who has always thought of himself as a \colorbox{lightyellow}{hopeless romantic}—but has never truly fallen in \colorbox{lightyellow}{love}. Focused on school and friendships, his life takes an unexpected turn when Grace Town joins his class. As they are paired to work on the school newspaper, Henry discovers the excitement, confusion, and intensity of \colorbox{lightyellow}{first love}—something any \colorbox{lightyellow}{young reader} navigating the ups and downs of \colorbox{lightyellow}{adolescence} can relate to. \\
\addlinespace
\textbf{PrefAlign} & \textit{Our Chemical Hearts} tells the story of Henry Page, a teenager who considers himself a \colorbox{lightyellow}{hopeless romantic} but has never fallen in \colorbox{lightyellow}{love}. Focused on school and his future, his world changes when Grace Town joins his class. She isn’t exactly the girl he expected, but as they are paired to work together on the school newspaper, Henry begins to experience the joys and challenges of \colorbox{lightyellow}{first love}, discovering how unpredictable and powerful emotions can be. \\
\addlinespace
\textbf{\approach} & \textit{Our Chemical Hearts} follows Henry Page, a \colorbox{lightyellow}{teenager} who has always thought of himself as a \colorbox{lightyellow}{hopeless romantic}—but has never truly fallen in \colorbox{lightyellow}{love}. 
\colorbox{lightpink}{Focused on school and his friends}, his life takes an unexpected turn when the \colorbox{lightblue}{intriguing and unpredictable Grace Town joins his class}. 
As they are paired to work on the school newspaper, Henry \colorbox{lightblue}{experiences the thrill, surprises, and intensity of first love}, along with the \colorbox{lightpink}{joys and challenges of navigating friendships and new experiences}—\colorbox{lightpink}{perfect for any} \colorbox{lightyellow}{young reader} who \colorbox{lightpink}{loves excitement, connection}. \\
\bottomrule
\end{tabular}
\vspace{-1mm}
\caption{Example personalized book descriptions for a simplified persona for select baselines. Our method better captures user-relevant themes compared to baselines. Highlighted texts show relevance to user's \interests, \values and \personality. We recommend viewing this table in full color.}
\label{tab:main_results_ex}
\end{table*}

\subsection{Experiment Setup}
\label{sec:exp_setup}

\noindent \textbf{Finetuning user-centric LLMs} We adapt \texttt{Llama-3.1-8B-Instruct} \cite{dubey2024llama} to user preferences using DPO with LoRA adapters \cite{hu2022lora}. We preference tune the model using the \texttt{TRL} library \cite{trl2023} with the AdamW optimizer \cite{loshchilov2017decoupled}, a $\beta$ value of 0.3, a learning rate of \(2 \times 10^{-5}\), weight decay of 0.01, a cosine learning rate schedule, and a warmup ratio of 0.05. Additional training details and prompt formats are detailed in \S\ref{app:our_method}.

\noindent \textbf{Personalization strategies} We benchmark against seven baselines\footnote{All methods use \texttt{Llama-3.1-8B-Instruct} unless noted.}, differing in how user information is integrated: (1) \textbf{Prompting}: (a) \textit{BaseRewrite} — rewrites the description to be more engaging (no personalization), (b) \textit{DemoBased} — appends user demographics (age, gender, location), (c) \textit{UserSummary} — prepends a GPT-4o–generated user profile summary, (d) \textit{LaMP} \cite{salemi2023lamp} — retrieves and prepends the user’s most relevant past reviews, (e) \textit{TriAgent} \cite{xiao2023personalized} — a three-stage pipeline combining \textit{BaseRewrite}, GPT-4o summaries, and customized edit instructions; (2) \textbf{SFT}: (a) \textit{UserSFT} — fine-tunes on (book description → user review) pairs with persona prompts encoding user demographics; (3) \textbf{Preference data}: (a) \textit{PrefAlign} — uses DPO using GPT-4o-generated aligned and misaligned book descriptions, produced solely from user demographics. Details on baseline implementations are in \S\ref{app:baselines}.

\noindent \textbf{Auto-evaluation metrics} To assess the quality of personalized descriptions, we utilize three personalization-measuring metrics: (1) \textit{Top-1 WinRate}: how often is a method's output preferred over all other methods, (2) \textit{Preference Gain}: how often is a method's output preferred over the original description, and (3) \textit{Interestingness Score}: how engaging/captivating does the user find the method's output (based on 5-point Likert score). We evaluate personalizations using LLM-as-a-judge \cite{zheng2023judgingllmasajudgemtbenchchatbot}. We provide GPT-4o with a detailed user persona, generated from the user’s complete set of written reviews and and available demographics. To ensure personalized content remains faithful to the original description, we also compute cosine similarity between SentenceBERT \cite{reimers2019sentence} embeddings of the original and generated texts. Prompt templates used for evaluation are included in \S\ref{app:metrics}.

\begin{table}[t]
\centering
\footnotesize
\setlength{\tabcolsep}{3pt} 
\begin{tabular}{lccc}
\toprule
\textbf{Method} & \textbf{Top-1 Win (\%)} & \textbf{Pref. Gain (\%)} & \textbf{Int. Score} \\
\midrule
Original & 7.78 & -- & 3.48 \\
TriAgent & 30.92 & 78.87 & 4.09 \\
\approach & \colorbox{lightgreen}{61.30} & \colorbox{lightgreen}{86.73} & \colorbox{lightgreen}{4.15} \\
\bottomrule
\end{tabular}
\caption{User Study personalization metrics (\textit{Top-1 Win Rate}, \textit{Preference Gain}, and \textit{Interestingness Score}) based on user rankings of book descriptions: original, LaMP, and \approach descriptions, aggregated over all users. Consistent with auto-evaluation metrics, we observe users prefer \approach generations over baseline generations and the original description.}
\label{tab:user_study_results}
\end{table}

\noindent\textbf{User study} We conduct a small study with 120 participants from diverse cultural backgrounds (30 users from each of USA, Brazil, Japan, and India). For each participant, we collect demographics, top book genres, values, and personality traits. We then train a customized model for these users and generate personalized book descriptions for each participant based on popular books in their preferred genres. Participants rank three descriptions of 10 unique books (original, \textit{TriAgent}, and \approach) based on how engaging and interesting they find each book description. Further details on our user study design are included in \S\ref{app:user_study}.

\begin{figure}[!t]
\centering
\includegraphics[width=1\columnwidth]{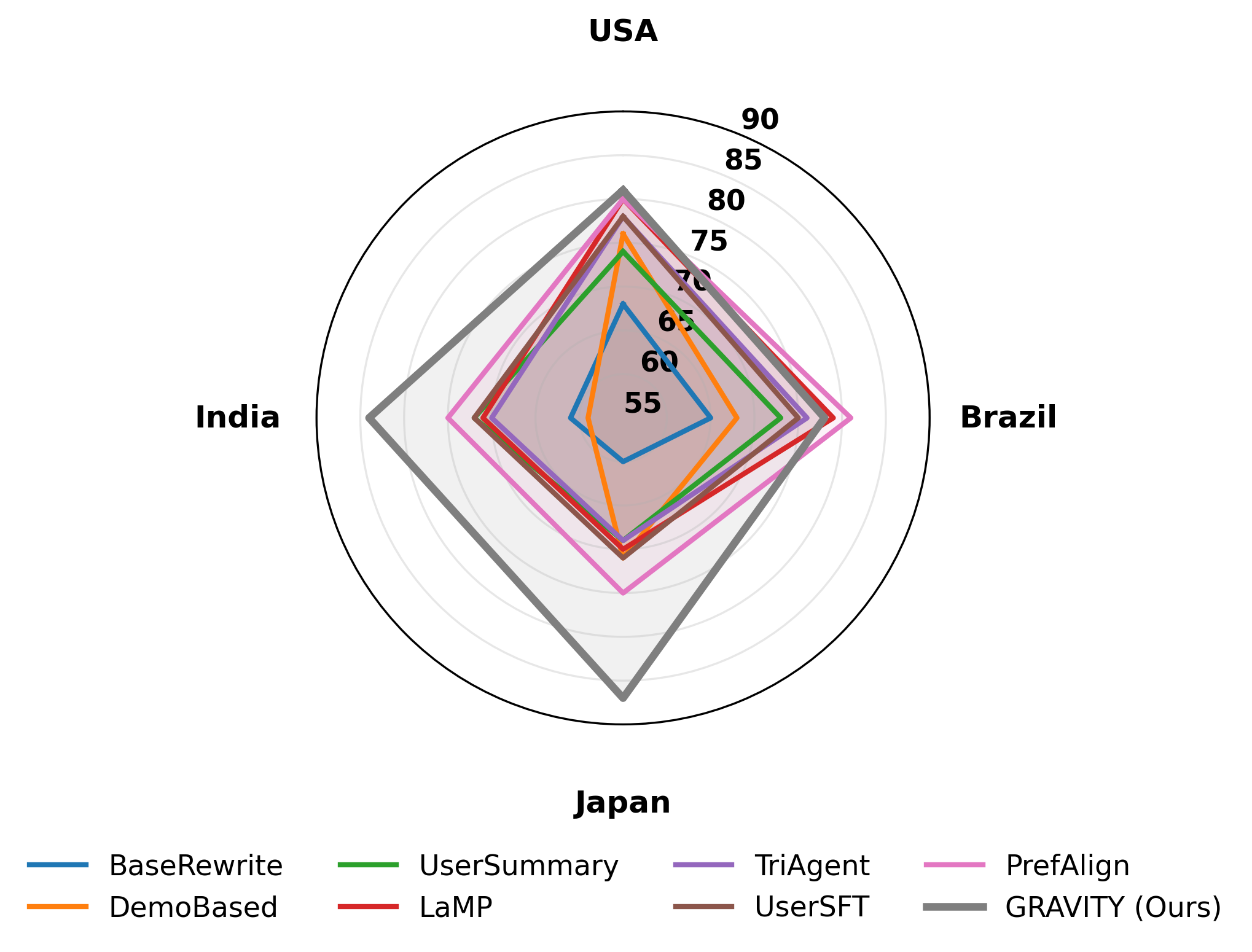}
\caption{Preference Gains (\%) for personalized generation methods across users from four diverse countries: USA, Brazil, Japan, and India. We observe that \approach yields consistently strong personalization metrics across the cultures, with gains in non-Western countries compared to current approaches (>10\% increase in preference gains).}
\label{fig:cultural_experiments}
\end{figure}  

\subsection{Personalization with \approach}
\label{sec:results}

\noindent \textbf{Comparisons against baselines} Table~\ref{tab:main_results} reports automatic evaluation metrics for all baselines and \approach. Our method achieves the highest win rates (4.5\% improvement) and preference gains (3.5\% improvement) compared to the strongest baselines. Interestingness scores are similarly high across baselines and \approach, with average ratings above 4, indicating that generated descriptions are engaging to users. Furthermore, more than 70\% of the content remains semantically consistent with the original description—comparable to other baselines—demonstrating that our method preserves the original content while personalizing it.

\noindent \textbf{User evaluations} Table \ref{tab:user_study_results} show \approach leads to generalizable performance gains across users from various countries and book categories. Compared to the best baseline, \textit{LaMP}, we observe that users tend to prefer book descriptions generated via \approach more than 61\% of the time. Moreover, users also tend to prefer \approach generations approximately 8\% more than original book descriptions. Users also tend to find descriptions generated through \approach as the most interesting. 

\begin{figure}[!t]
\centering
\includegraphics[width=\columnwidth]{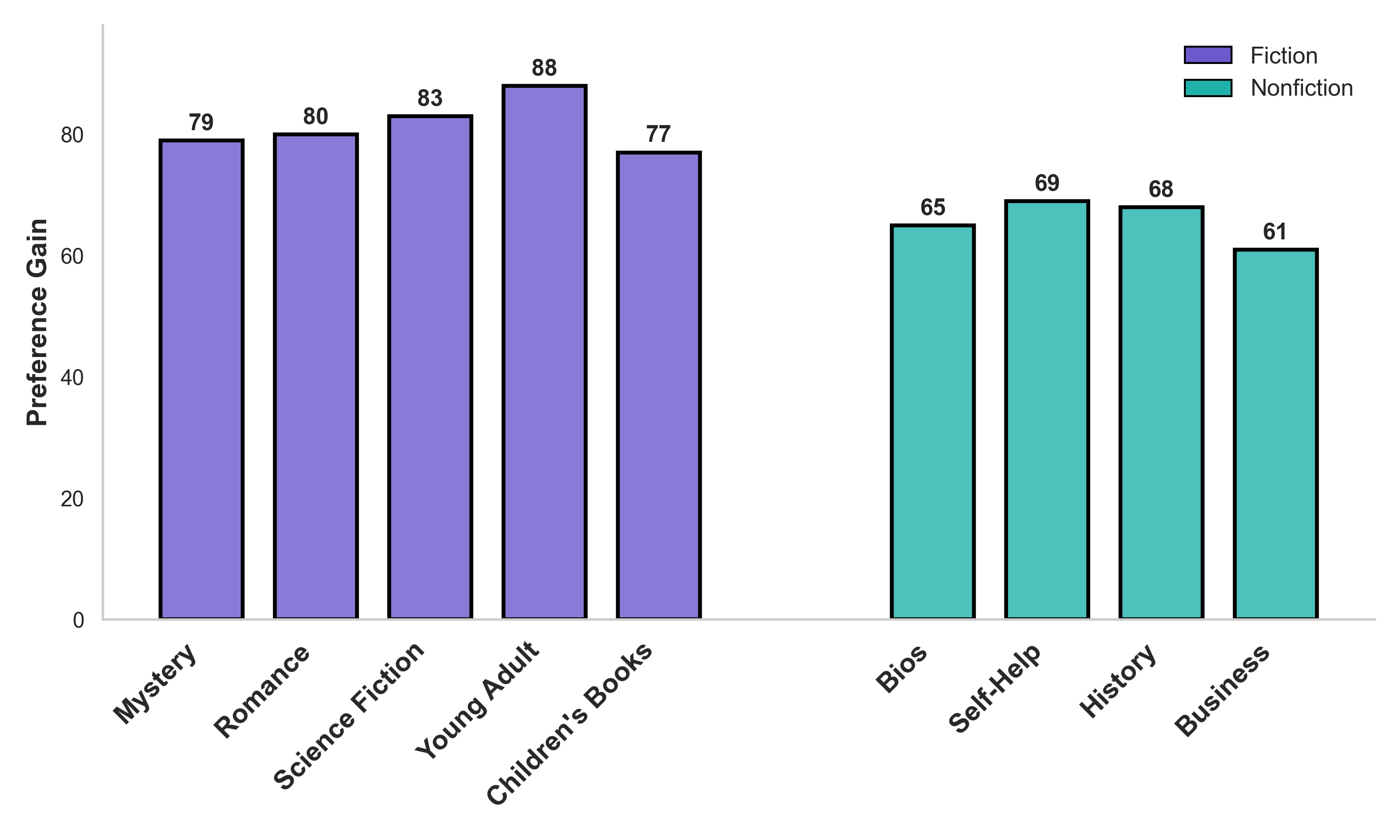}
\caption{Preference gains (\%) across nine book genres (similar categories, clustered into fiction and non-fiction) using \approach. We find that most fiction books have much higher preference gains (> 70\%) as compared to non-fiction books. ($\approx$ 65\%).}
\label{fig:genre_experiments}
\end{figure} 

\noindent \textbf{Cross-Cultural Personalization}
To evaluate robustness across diverse populations, we report preference gains for users from the USA, Brazil, Japan, and India (Figure~\ref{fig:cultural_experiments}). Results are averaged across each user group for the six baselines and our approach, \approach. We find that \approach achieves consistent improvements in all regions, with higher preferences in almost all countries. Particularly, we observe large yields (>10\%) increase) in non-Western contexts such as Japan. Although Brazilians may prefer \textit{PrefAlign} generations approximately 3\% more than \approach, we still observe comparable performances across personalization metrics (see Table \ref{tab:brazil_results}) with \approach. These results indicate that tailoring preference data to capture users’ cultural value systems and personality traits may substantially improve alignment with user's engagement and preferences. 

\begin{table}[!t]
\centering
\footnotesize
\setlength{\tabcolsep}{8pt}
\renewcommand{\arraystretch}{1.2}

\begin{tabular}{l S[table-format=-2.2]}
\toprule
\textbf{Feature} & {\textbf{Preference Gain (\%)}} \\
\midrule
All (original \approach) & 86.73 \\ 
\midrule 
- User Interests & -10.34** \\ 
\quad - Category & -6.23** \\ 
\quad - Summary & -4.38* \\ 
- Values \& Beliefs & -8.92** \\
- Personality & -8.48** \\ 
\quad - Openness & -5.45** \\ 
\quad - Conscientiousness & -2.35* \\
\quad - Extraversion & -6.03** \\
\quad - Agreeableness & -1.32 \\
\quad - Neuroticism & -0.98 \\
\bottomrule
\end{tabular}

\caption{Ablation studies showing the impact of different user preference data (interests, values, and personality traits) on personalization performance, measured via decreases in preference gains. Subsets of Values \& Beliefs data are not removed due to small size (2K–3K pairs per user). Statistical significance is assessed using paired Wilcoxon signed-rank tests: * indicates $p < 0.05$, and ** indicates $p < 0.01$.}
\label{tab:ablations}
\end{table}

\noindent \textbf{Personalization Across Genres} We further analyze how personalization effectiveness varies across book genres. While our method consistently improves over baselines across both fiction and nonfiction, we observe substantially higher gains for fiction titles (see Figure \ref{fig:genre_experiments}\footnote{Although the Amazon Reviews dataset has several fine-grained labels for book genres, in this analysis, we focus on a set of nine hierarchical book categories/themes.}); we see preference gainst above 75\% for fiction books but gains closer to 65\% for non-fiction books. Our findings align with prior research which show that fiction readers tend to be more engaged with narrative elements—such as character development, emotional arcs, and imaginative settings—which naturally connect with readers’ personal values, beliefs, and personality traits \cite{bal2013does, goyal2024systematic}. Thus, incorporating these data as training signals into user-centric models allows for better personalization. In contrast, nonfiction reading is primarily fact-oriented, emphasizing clarity, accuracy, and logical structure, which limits the potential for personalization, resulting in smaller improvements through \approach.

\subsection{Effects of Custom Preference Data}
\label{sec:ablations}

To understand the contribution of different components of user preference data to personalization performance, we perform an ablation study. Specifically, we remove each data subset: \interests, \values, and \personality, and measure the resulting change in preference gains compared to the full model (Table~\ref{tab:ablations}) (using GPT-4o judge, see \S\ref{app:metrics} for prompt details).

We observe that removing any preference dataset component leads to lead to statistically significant drops in preference gains, highlighting the importance of each type of user information. Notably, apart from \interests, \values have the largest impact, with decreases of almost 9\%, while individual \personality such as \textit{Openness} and \textit{Extraversion} also contribute substantially. Some traits, like \textit{Agreeableness} and \textit{Neuroticism}, have a smaller effect, suggesting that their influence on personalization may be more subtle. 

\subsection{SFT vs. DPO}
\label{sec:sft_results}

While DPO enables models to directly learn user preferences and stylistic nuances, supervised finetuning (SFT) offers a simpler, more stable alternative that is less sensitive to hyperparameter choices and requires less training resources. To study this trade-off, we evaluate a setting where models are taught about a user’s preferred book genres, values, and personality traits through SFT. For each user, we construct binary training examples of the form: given demographic attributes and a scenario, the user would most likely \underline{$A$}, where $A$ corresponds to the chosen response from each preference pair.  

As in our original approach, we train \texttt{Llama-3.1-8B-Instruct}, using LoRA for 5 epochs using a learning rate of $10^{-4}$ with GPT-4o as our evaluation judge. Table~\ref{tab:sft_metrics} reports results across baselines and \approach: SFT and DPO.  

\begin{table}[t]
\centering
\scriptsize
\setlength{\tabcolsep}{3pt} 
\begin{tabular}{lccc}
\toprule
\textbf{Method} & \textbf{Top-1 Win (\%)} & \textbf{Pref. Gain (\%)} & \textbf{Int. Score} \\
\midrule
BaseRewrite & 1.0 & 63.5 & 3.81 \\
DemoBased & 3.0 & 68.5 & 3.69 \\
UserSummary & 4.5 & 72.0 & 3.98 \\
LaMP & 4.75 & 75.5 & \colorbox{lightgreen}{4.16} \\
TriAgent & 4.25 & 73.25 & 3.95 \\
UserSFT & 17.0 & 74.25 & 4.04 \\
PrefAlign & 20.75 & 79.0 & 4.24 \\ 
\approach (SFT) & 20.25 & 72.58 & 3.98 \\
\approach (DPO) & \colorbox{lightgreen}{24.5} & \colorbox{lightgreen}{82.67} & 4.10 \\
\bottomrule
\end{tabular}
\caption{Personalization metrics with SFT on \approach data. SFT improves engagement (comparable to \textit{PrefAlign} and \textit{UserSFT} method results), while DPO achieves the strongest gains in win rate, preference, and interestingness.}
\label{tab:sft_metrics}
\end{table}

Overall, our findings suggest that incorporating user-specific data (interests, values, and personality) enhances personalization across methods. However, DPO more effectively captures fine-grained preferences, leading to the highest improvements in alignment and engagement for win rate, preference gain, and user interestingness.

\section{Conclusion}

In this work, we presented \approach, a framework for generating personalized text by leveraging synthetic, profile-grounded preference data derived from demographics, interests, values, beliefs, and personality traits. By integrating psychological and cultural frameworks, our approach enables LLMs to align content with diverse user profiles and capture meaningful variation in individual preferences.

Applied to personalized Amazon book descriptions, \approach demonstrates that scenario-driven preference data can effectively guide model behavior, with deeper attributes such as values and personality providing measurable gains beyond demographics and interests. These results highlight the potential of structured synthetic preference data as a scalable and interpretable approach to user-centric LLM personalization.

\section{Limitations}

In this work, we approximate user attributes in order to build a synthetic preference dataset. While this approach reduces the need for costly annotation, it may not guarantee exact alignment with an individual’s intrinsic value systems, cultural beliefs, or personality traits. Prior work has shown that such attributes can be estimated through model prompting and training, but we acknowledge that profile inference introduces uncertainty and potential noise. To partly mitigate this, we conducted a user study (see \S\ref{app:user_study}) to validate whether personalized generations align with the participants’ self-reported preferences and profile attributes. 

Another limitation lies in cross-cultural generalization. Although we ground \approach in established cultural psychology frameworks such as Hofstede’s dimensions, Schwartz’s values, and the World Values Survey, these instruments may not fully capture the richness and evolving nature of cultural identity. Our results show meaningful gains across different cultural groups, but personalization may still miss subtle or context-dependent user preferences that fall outside of these frameworks.

Finally, this work focuses on book description personalization using Amazon reviews as a controlled testbed for exploring the effectiveness of \approach. The main goal of this work is to demonstrate the feasibility of using synthetic, profile-grounded preference data for personalization, but future work can test its applicability across broader datasets and domains, such as news, healthcare, and education, where personalization challenges may differ.

\section*{Acknowledgments} 
 We would like to thank researchers from USC LIME and HUMANS labs for their continuous feedback in the formulation and setting of our work. This research was supported in part by the NSF, under Award Number 2331722. Priyanka Dey was also supported by the 2025 USC Capital One Fellowship. 

\bibliography{acl_latex}

@article{goyal2024systematic,
  title={A systematic review of synthetic data generation techniques using generative AI},
  author={Goyal, Mandeep and Mahmoud, Qusay H},
  journal={Electronics},
  volume={13},
  number={17},
  pages={3509},
  year={2024},
  publisher={MDPI}
}

@article{shao2024ulmrec,
  title={ULMRec: User-centric large language model for sequential recommendation},
  author={Shao, Minglai and Huang, Hua and Peng, Qiyao and Liu, Hongtao},
  journal={arXiv preprint arXiv:2412.05543},
  year={2024}
}

@article{zhong2024panacea,
  title={Panacea: Pareto alignment via preference adaptation for llms},
  author={Zhong, Yifan and Ma, Chengdong and Zhang, Xiaoyuan and Yang, Ziran and Chen, Haojun and Zhang, Qingfu and Qi, Siyuan and Yang, Yaodong},
  journal={Advances in Neural Information Processing Systems},
  volume={37},
  pages={75522--75558},
  year={2024}
}

@article{shenfeld2025language,
  title={Language model personalization via reward factorization},
  author={Shenfeld, Idan and Faltings, Felix and Agrawal, Pulkit and Pacchiano, Aldo},
  journal={arXiv preprint arXiv:2503.06358},
  year={2025}
}

@article{bismay2024reasoningrec,
  title={Reasoningrec: Bridging personalized recommendations and human-interpretable explanations through llm reasoning},
  author={Bismay, Millennium and Dong, Xiangjue and Caverlee, James},
  journal={arXiv preprint arXiv:2410.23180},
  year={2024}
}

@inproceedings{acharya2023llm,
  title={Llm based generation of item-description for recommendation system},
  author={Acharya, Arkadeep and Singh, Brijraj and Onoe, Naoyuki},
  booktitle={Proceedings of the 17th ACM conference on recommender systems},
  pages={1204--1207},
  year={2023}
}

@article{li2024personalized,
  title={Personalized language modeling from personalized human feedback},
  author={Li, Xinyu and Zhou, Ruiyang and Lipton, Zachary C and Leqi, Liu},
  journal={arXiv preprint arXiv:2402.05133},
  year={2024}
}

@article{peng2024review,
  title={Review-llm: Harnessing large language models for personalized review generation},
  author={Peng, Qiyao and Liu, Hongtao and Xu, Hongyan and Yang, Qing and Shao, Minglai and Wang, Wenjun},
  journal={arXiv preprint arXiv:2407.07487},
  year={2024}
}

@article{jang2023personalized,
  title={Personalized soups: Personalized large language model alignment via post-hoc parameter merging},
  author={Jang, Joel and Kim, Seungone and Lin, Bill Yuchen and Wang, Yizhong and Hessel, Jack and Zettlemoyer, Luke and Hajishirzi, Hannaneh and Choi, Yejin and Ammanabrolu, Prithviraj},
  journal={arXiv preprint arXiv:2310.11564},
  year={2023}
}

@article{jiang2023personallm,
  title={PersonaLLM: Investigating the ability of large language models to express personality traits},
  author={Jiang, Hang and Zhang, Xiajie and Cao, Xubo and Breazeal, Cynthia and Roy, Deb and Kabbara, Jad},
  journal={arXiv preprint arXiv:2305.02547},
  year={2023}
}

@article{dey2025can,
  title={Can LLMs Express Personality Across Cultures? Introducing CulturalPersonas for Evaluating Trait Alignment},
  author={Dey, Priyanka and Khanter, Yugal and Bothra, Aayush and Zhao, Jieyu and Ferrara, Emilio},
  journal={arXiv preprint arXiv:2506.05670},
  year={2025}
}

@article{joshi2025improving,
  title={Improving llm personas via rationalization with psychological scaffolds},
  author={Joshi, Brihi and Ren, Xiang and Swayamdipta, Swabha and Koncel-Kedziorski, Rik and Paek, Tim},
  journal={arXiv e-prints},
  pages={arXiv--2504},
  year={2025}
}

@article{liu2025can,
  title={Can LLMs Grasp Implicit Cultural Values? Benchmarking LLMs' Metacognitive Cultural Intelligence with CQ-Bench},
  author={Liu, Ziyi and Dey, Priyanka and Zhao, Zhenyu and Huang, Jen-tse and Gupta, Rahul and Liu, Yang and Zhao, Jieyu},
  journal={arXiv preprint arXiv:2504.01127},
  year={2025}
}

@misc{zheng2025tuningfreellmbuildstrong,
      title={Tuning-Free LLM Can Build A Strong Recommender Under Sparse Connectivity And Knowledge Gap Via Extracting Intent}, 
      author={Wenqing Zheng and Noah Fatsi and Daniel Barcklow and Dmitri Kalaev and Steven Yao and Owen Reinert and C. Bayan Bruss and Daniele Rosa},
      year={2025},
      eprint={2505.10900},
      archivePrefix={arXiv},
      primaryClass={cs.IR},
      url={https://arxiv.org/abs/2505.10900}, 
}

@article{donnellan2006mini,
  title={The mini-IPIP scales: tiny-yet-effective measures of the Big Five factors of personality.},
  author={Donnellan, M Brent and Oswald, Frederick L and Baird, Brendan M and Lucas, Richard E},
  journal={Psychological assessment},
  volume={18},
  number={2},
  pages={192},
  year={2006},
  publisher={American Psychological Association}
}

@article{Douze2024TheFL,
  title={The Faiss library},
  author={Matthijs Douze and Alexandr Guzhva and Chengqi Deng and Jeff Johnson and Gergely Szilvasy and Pierre-Emmanuel Mazare and Maria Lomeli and Lucas Hosseini and Herv{\'e} J{\'e}gou},
  journal={arXiv preprint arXiv:2401.08281},
  year={2024}
}

@misc{rajbhandari2020zeromemoryoptimizationstraining,
      title={ZeRO: Memory Optimizations Toward Training Trillion Parameter Models}, 
      author={Samyam Rajbhandari and Jeff Rasley and Olatunji Ruwase and Yuxiong He},
      year={2020},
      eprint={1910.02054},
      archivePrefix={arXiv},
      primaryClass={cs.LG},
      url={https://arxiv.org/abs/1910.02054}, 
}

@misc{yunusov2024mirrorstoriesreflectingdiversitypersonalized,
      title={MirrorStories: Reflecting Diversity through Personalized Narrative Generation with Large Language Models}, 
      author={Sarfaroz Yunusov and Hamza Sidat and Ali Emami},
      year={2024},
      eprint={2409.13935},
      archivePrefix={arXiv},
      primaryClass={cs.CL},
      url={https://arxiv.org/abs/2409.13935}, 
}

@inproceedings{cai-etal-2023-generating,
    title = "Generating User-Engaging News Headlines",
    author = "Cai, Pengshan  and
      Song, Kaiqiang  and
      Cho, Sangwoo  and
      Wang, Hongwei  and
      Wang, Xiaoyang  and
      Yu, Hong  and
      Liu, Fei  and
      Yu, Dong",
    editor = "Rogers, Anna  and
      Boyd-Graber, Jordan  and
      Okazaki, Naoaki",
    booktitle = "Proceedings of the 61st Annual Meeting of the Association for Computational Linguistics (Volume 1: Long Papers)",
    month = jul,
    year = "2023",
    address = "Toronto, Canada",
    publisher = "Association for Computational Linguistics",
    url = "https://aclanthology.org/2023.acl-long.183/",
    doi = "10.18653/v1/2023.acl-long.183",
    pages = "3265--3280"
}

@inproceedings{moorjani-etal-2022-audience,
    title = "Audience-Centric Natural Language Generation via Style Infusion",
    author = "Moorjani, Samraj  and
      Krishnan, Adit  and
      Sundaram, Hari  and
      Maslowska, Ewa  and
      Sankar, Aravind",
    editor = "Goldberg, Yoav  and
      Kozareva, Zornitsa  and
      Zhang, Yue",
    booktitle = "Findings of the Association for Computational Linguistics: EMNLP 2022",
    month = dec,
    year = "2022",
    address = "Abu Dhabi, United Arab Emirates",
    publisher = "Association for Computational Linguistics",
    url = "https://aclanthology.org/2022.findings-emnlp.138/",
    doi = "10.18653/v1/2022.findings-emnlp.138",
    pages = "1919--1932"
}

@inproceedings{zhang-etal-2025-personalized,
    title = "Personalized Text Generation with Contrastive Activation Steering",
    author = "Zhang, Jinghao  and
      Liu, Yuting  and
      Wang, Wenjie  and
      Liu, Qiang  and
      Wu, Shu  and
      Wang, Liang  and
      Chua, Tat-Seng",
    editor = "Che, Wanxiang  and
      Nabende, Joyce  and
      Shutova, Ekaterina  and
      Pilehvar, Mohammad Taher",
    booktitle = "Proceedings of the 63rd Annual Meeting of the Association for Computational Linguistics (Volume 1: Long Papers)",
    month = jul,
    year = "2025",
    address = "Vienna, Austria",
    publisher = "Association for Computational Linguistics",
    url = "https://aclanthology.org/2025.acl-long.353/",
    doi = "10.18653/v1/2025.acl-long.353",
    pages = "7128--7141",
    ISBN = "979-8-89176-251-0"
}

@article{matz2024potential,
  title={The potential of generative AI for personalized persuasion at scale},
  author={Matz, Sandra C and Teeny, Jacob D and Vaid, Sumer S and Peters, Heinrich and Harari, Gabriella M and Cerf, Moran},
  journal={Scientific Reports},
  volume={14},
  number={1},
  pages={4692},
  year={2024},
  publisher={Nature Publishing Group UK London}
}

@article{nguyen2018user,
  title={User personality and user satisfaction with recommender systems},
  author={Nguyen, Tien T and Maxwell Harper, F and Terveen, Loren and Konstan, Joseph A},
  journal={Information systems frontiers},
  volume={20},
  number={6},
  pages={1173--1189},
  year={2018},
  publisher={Springer}
}

@article{yang2023palr,
  title={Palr: Personalization aware llms for recommendation},
  author={Yang, Fan and Chen, Zheng and Jiang, Ziyan and Cho, Eunah and Huang, Xiaojiang and Lu, Yanbin},
  journal={arXiv preprint arXiv:2305.07622},
  year={2023}
}

@inproceedings{wozniak2024personalized,
  title={Personalized large language models},
  author={Wo{\'z}niak, Stanis{\l}aw and Koptyra, Bart{\l}omiej and Janz, Arkadiusz and Kazienko, Przemys{\l}aw and Koco{\'n}, Jan},
  booktitle={2024 IEEE International Conference on Data Mining Workshops (ICDMW)},
  pages={511--520},
  year={2024},
  organization={IEEE}
}

@article{rafailov2023direct,
  title={Direct preference optimization: Your language model is secretly a reward model},
  author={Rafailov, Rafael and Sharma, Archit and Mitchell, Eric and Manning, Christopher D and Ermon, Stefano and Finn, Chelsea},
  journal={Advances in neural information processing systems},
  volume={36},
  pages={53728--53741},
  year={2023}
}

@incollection{goldberg2013alternative,
  title={An alternative “description of personality”: The Big-Five factor structure},
  author={Goldberg, Lewis R},
  booktitle={Personality and personality disorders},
  pages={34--47},
  year={2013},
  publisher={Routledge}
}

@MISC{Haerpfer2024-jz,
  title     = "World Values Survey wave 7 (2017-2022) cross-national data-set",
  author    = "Haerpfer, Christian and Inglehart, Ronald and Moreno, Alejandro
               and Welzel, Christian and Kizilova, Kseniya and Diez-Medrano,
               Jaime and Lagos, Marta and Norris, Pippa and Ponarin, Eduard and
               Puranen, Bi",
  publisher = "World Values Survey Association",
  year      =  2024
}

@article{schwartz2012overview,
  title={An overview of the Schwartz theory of basic values},
  author={Schwartz, Shalom H},
  journal={Online readings in Psychology and Culture},
  volume={2},
  number={1},
  pages={11},
  year={2012}
}

@article{hofstede1983national,
  title={National cultures in four dimensions: A research-based theory of cultural differences among nations},
  author={Hofstede, Geert},
  journal={International studies of management \& organization},
  volume={13},
  number={1-2},
  pages={46--74},
  year={1983},
  publisher={Taylor \& Francis}
}

@inproceedings{zhang2024p4,
  title={P4: Plug-and-Play Discrete Prompting for Large Language Models Personalization},
  author={Zhang, Yuansen and Wang, Xiao and Chen, Tianze and Fu, Jiayi and Gui, Tao and Zhang, Qi},
  booktitle={Findings of the Association for Computational Linguistics ACL 2024},
  pages={9129--9144},
  year={2024}
}

@article{lyu2023llm,
  title={Llm-rec: Personalized recommendation via prompting large language models},
  author={Lyu, Hanjia and Jiang, Song and Zeng, Hanqing and Xia, Yinglong and Wang, Qifan and Zhang, Si and Chen, Ren and Leung, Christopher and Tang, Jiajie and Luo, Jiebo},
  journal={arXiv preprint arXiv:2307.15780},
  year={2023}
}

@article{lee2024aligning,
  title={Aligning to thousands of preferences via system message generalization},
  author={Lee, Seongyun and Park, Sue Hyun and Kim, Seungone and Seo, Minjoon},
  journal={Advances in Neural Information Processing Systems},
  volume={37},
  pages={73783--73829},
  year={2024}
}

@article{poddar2024personalizing,
  title={Personalizing reinforcement learning from human feedback with variational preference learning},
  author={Poddar, Sriyash and Wan, Yanming and Ivison, Hamish and Gupta, Abhishek and Jaques, Natasha},
  journal={Advances in Neural Information Processing Systems},
  volume={37},
  pages={52516--52544},
  year={2024}
}

@article{kirk2024prism,
  title={The PRISM alignment dataset: What participatory, representative and individualised human feedback reveals about the subjective and multicultural alignment of large language models},
  author={Kirk, Hannah Rose and Whitefield, Alexander and Rottger, Paul and Bean, Andrew M and Margatina, Katerina and Mosquera-Gomez, Rafael and Ciro, Juan and Bartolo, Max and Williams, Adina and He, He and others},
  journal={Advances in Neural Information Processing Systems},
  volume={37},
  pages={105236--105344},
  year={2024}
}

@article{zhang2024personalization,
  title={Personalization of large language models: A survey},
  author={Zhang, Zhehao and Rossi, Ryan A and Kveton, Branislav and Shao, Yijia and Yang, Diyi and Zamani, Hamed and Dernoncourt, Franck and Barrow, Joe and Yu, Tong and Kim, Sungchul and others},
  journal={arXiv preprint arXiv:2411.00027},
  year={2024}
}

@article{chen2024persona,
  title={From persona to personalization: A survey on role-playing language agents},
  author={Chen, Jiangjie and Wang, Xintao and Xu, Rui and Yuan, Siyu and Zhang, Yikai and Shi, Wei and Xie, Jian and Li, Shuang and Yang, Ruihan and Zhu, Tinghui and others},
  journal={arXiv preprint arXiv:2404.18231},
  year={2024}
}

@article{bal2013does,
  title={How does fiction reading influence empathy? An experimental investigation on the role of emotional transportation},
  author={Bal, P Matthijs and Veltkamp, Martijn},
  journal={PloS one},
  volume={8},
  number={1},
  pages={e55341},
  year={2013},
  publisher={Public Library of Science San Francisco, USA}
}

@article{hou2024bridging,
  title={Bridging Language and Items for Retrieval and Recommendation},
  author={Hou, Yupeng and Li, Jiacheng and He, Zhankui and Yan, An and Chen, Xiusi and McAuley, Julian},
  journal={arXiv preprint arXiv:2403.03952},
  year={2024}
}

@misc{kaantureyyen2023deberta_gender,
  title        = {kaantureyyen/deberta-blog-authorship-corpus-gender: DeBERTaV3 (small) fine-tuned on the Blog Authorship Corpus for gender classification},
  author       = {Tureyyen, Kaan},
  howpublished = {\url{https://huggingface.co/kaantureyyen/deberta-blog-authorship-corpus-gender}},
  year         = {2023}
}

@article{dubey2024llama,
  title={The llama 3 herd of models},
  author={Dubey, Abhimanyu and Jauhri, Abhinav and Pandey, Abhinav and Kadian, Abhishek and Al-Dahle, Ahmad and Letman, Aiesha and Mathur, Akhil and Schelten, Alan and Yang, Amy and Fan, Angela and others},
  journal={arXiv e-prints},
  pages={arXiv--2407},
  year={2024}
}

@article{hu2022lora,
  title={Lora: Low-rank adaptation of large language models.},
  author={Hu, Edward J and Shen, Yelong and Wallis, Phillip and Allen-Zhu, Zeyuan and Li, Yuanzhi and Wang, Shean and Wang, Lu and Chen, Weizhu and others},
  journal={ICLR},
  volume={1},
  number={2},
  pages={3},
  year={2022}
}

@misc{trl2023,
  author = {CarperAI},
  title = {TRL: Transformer Reinforcement Learning},
  year = {2023},
  url = {https://huggingface.co/docs/trl/en/index}
}

@inproceedings{santurkar2023whose,
  title={Whose opinions do language models reflect?},
  author={Santurkar, Shibani and Durmus, Esin and Ladhak, Faisal and Lee, Cinoo and Liang, Percy and Hashimoto, Tatsunori},
  booktitle={International Conference on Machine Learning},
  pages={29971--30004},
  year={2023},
  organization={PMLR}
}

@article{hwang2023aligning,
  title={Aligning language models to user opinions},
  author={Hwang, EunJeong and Majumder, Bodhisattwa Prasad and Tandon, Niket},
  journal={arXiv preprint arXiv:2305.14929},
  year={2023}
}

@inproceedings{huang2023learning,
  title={Learning preference model for llms via automatic preference data generation},
  author={Huang, Shijia and Zhao, Jianqiao and Li, Yanyang and Wang, Liwei},
  booktitle={Proceedings of the 2023 Conference on Empirical Methods in Natural Language Processing},
  pages={9187--9199},
  year={2023}
}

@misc{singh2025fspofewshotpreferenceoptimization,
      title={FSPO: Few-Shot Preference Optimization of Synthetic Preference Data in LLMs Elicits Effective Personalization to Real Users}, 
      author={Anikait Singh and Sheryl Hsu and Kyle Hsu and Eric Mitchell and Stefano Ermon and Tatsunori Hashimoto and Archit Sharma and Chelsea Finn},
      year={2025},
      eprint={2502.19312},
      archivePrefix={arXiv},
      primaryClass={cs.LG},
      url={https://arxiv.org/abs/2502.19312}, 
}

@article{loshchilov2017decoupled,
  title={Decoupled weight decay regularization},
  author={Loshchilov, Ilya and Hutter, Frank},
  journal={arXiv preprint arXiv:1711.05101},
  year={2017}
}

@misc{zheng2023judgingllmasajudgemtbenchchatbot,
      title={Judging LLM-as-a-Judge with MT-Bench and Chatbot Arena}, 
      author={Lianmin Zheng and Wei-Lin Chiang and Ying Sheng and Siyuan Zhuang and Zhanghao Wu and Yonghao Zhuang and Zi Lin and Zhuohan Li and Dacheng Li and Eric P. Xing and Hao Zhang and Joseph E. Gonzalez and Ion Stoica},
      year={2023},
      eprint={2306.05685},
      archivePrefix={arXiv},
      primaryClass={cs.CL},
      url={https://arxiv.org/abs/2306.05685}, 
}

@article{wang2024continuous,
  title={Continuous output personality detection models via mixed strategy training},
  author={Wang, Rong and Sun, Kun},
  journal={arXiv preprint arXiv:2406.16223},
  year={2024}
}

@misc{kaantureyyen2023deberta_age,
  title        = {kaantureyyen/deberta-blog-authorship-corpus-age: DeBERTaV3 (small) fine-tuned on the Blog Authorship Corpus for age classification},
  author       = {Tureyyen, Kaan},
  howpublished = {\url{https://huggingface.co/kaantureyyen/deberta-blog-authorship-corpus-age}},
  year         = {2023}
}

@article{xiao2023personalized,
  title={Personalized abstractive summarization by tri-agent generation pipeline},
  author={Xiao, Wen and Xie, Yujia and Carenini, Giuseppe and He, Pengcheng},
  journal={arXiv preprint arXiv:2305.02483},
  year={2023}
}

@article{song2023general,
  title={General then Personal: Decoupling and Pre-training for Personalized Headline Generation},
  author={Song, Yun-Zhu and Chen, Yi-Syuan and Wang, Lu and Shuai, Hong-Han},
  journal={Transactions of the Association for Computational Linguistics},
  volume={11},
  pages={1588--1607},
  year={2023},
  publisher={MIT Press One Broadway, 12th Floor, Cambridge, Massachusetts 02142, USA~…}
}

@article{tseng2024two,
  title={Two tales of persona in llms: A survey of role-playing and personalization},
  author={Tseng, Yu-Min and Huang, Yu-Chao and Hsiao, Teng-Yun and Chen, Wei-Lin and Huang, Chao-Wei and Meng, Yu and Chen, Yun-Nung},
  journal={arXiv preprint arXiv:2406.01171},
  year={2024}
}

@article{salemi2023lamp,
  title={Lamp: When large language models meet personalization},
  author={Salemi, Alireza and Mysore, Sheshera and Bendersky, Michael and Zamani, Hamed},
  journal={arXiv preprint arXiv:2304.11406},
  year={2023}
}

@article{Lee2024TRait,
  title={Do LLMs Have Distinct and Consistent Personality? TRAIT: Personality Testset designed for LLMs with Psychometrics},
  author={Seungbeen Lee and Seungwon Lim and Seungju Han and Giyeong Oh and Hyungjoo Chae and Jiwan Chung and Minju Kim and Beong-woo Kwak and Yeonsoo Lee and Dongha Lee and Jinyoung Yeo and Youngjae Yu},
  journal={arXiv preprint arXiv:2406.14703},
  year={2024}
}

@article{li2024big5chat,
  title   = {Big5-Chat: Shaping LLM Personalities Through Training on Human-Grounded Data},
  author  = {Li, Wenkai and Liu, Jiarui and Liu, Andy and Zhou, Xuhui and Diab, Mona T. and Sap, Maarten},
  journal = {arXiv preprint arXiv:2410.16491},
  year    = {2024}
}

@article{reimers2019sentence,
  title={Sentence-bert: Sentence embeddings using siamese bert-networks},
  author={Reimers, Nils and Gurevych, Iryna},
  journal={arXiv preprint arXiv:1908.10084},
  year={2019}
}

\clearpage

\appendix
\label{sec:appendix}

\section{Dataset}
\label{app:dataset}

As a case study for evaluating \approach, we utilize the Amazon Reviews Dataset collected in 2023 and contains over 571.54M reviews spanning May. 1996 to Sep. 2023. This dataset consists of product reviews from a wide range of categories, including books, clothing, and electronics. For our study, we focus on users who are avid readers and presenting them with personalized book descriptions. The original dataset consists of approximately 22.5M total book reviews from over 5M users. In our study, we select a small set of users (400) from 4 varying countries: United States, Brazil, India, and Japan and those who have rated at least 50 books. In Table \ref{tab:user_details}, we present more details on the randomly selected users. 

\begin{table}
\centering
\scriptsize
\setlength{\tabcolsep}{3pt} 
\begin{tabularx}{\linewidth}{l c X c} 
\toprule
\textbf{Country} & \textbf{Total Users} & \textbf{Top Book Genres} & \textbf{Avg. \# of Reviews} \\
\midrule
USA & 100 & Young Adult, Romance, History, Business & 79 \\
Brazil & 100 & Historical Fiction, Philosophy, Art History, Motivational & 61 \\
India & 100 & Thriller/Mystery, Historical Fiction, Biographies/Memoirs, Self-Improvement & 56 \\
Japan & 100 & Poetry, Folk literature, Engineering, Biographies/Memoirs & 53 \\
\bottomrule
\end{tabularx}
\caption{Details of users randomly selected for our case study of book description personalization (auto-evaluation).}
\label{tab:user_details}
\end{table}

\section{Custom Preference Generation}

In this section, we provide additional details for custom preference generation for each user. 

\interests We construct user interest data from each user’s top book genres. Across 400 users, we identify 382 unique categories (e.g., Mythology, Classical Literature, Self-Improvement). To identify the most distinct categories, we embed all category names using \texttt{all-MiniLM-L6-v2} (a Sentence-BERT model) and compute pairwise cosine similarities. For each user, we compare their top category with all others and select the three least similar (i.e., most dissimilar) categories. We then form \textit{Category} pairs by combining each user’s top genres with these inferred distinct categories. To construct \textit{Summary} pairs, we retrieve three highly rated books from both the user’s top and distinct categories, and pair them using their original book descriptions. This process yields roughly 90–240 \textit{Interests} pairs per user.

\values To construct values preference data, we focus on generating a set of 150 seed statements from various cultural frameworks on a wide variety of topics including culture, society, ethics, religion, politics, and morals. Table \ref{tab:add_seeds} consists of example statements generated from the various frameworks and topics. 

\begin{table*}[t]
\centering
\scriptsize
\setlength{\tabcolsep}{6pt} 
\begin{tabular}{>{\raggedright\arraybackslash}p{4cm} >{\raggedright\arraybackslash}p{9cm} > {\raggedright\arraybackslash}p{2cm}}
\toprule
\textbf{Source} & 
\textbf{Seed Statement} & 
\textbf{Topic}
\\
\addlinespace
\midrule

\multirow{5}{4cm}{\textbf{Hofstede's Cultural Dimensions} \newline \cite{hofstede1983national}} 
& Competition and achievement are more valued than cooperation and care. & Culture \\
& It is better to rely on clear rules and traditions than to face unpredictable situations without guidance. & Ethics \\
& A person’s identity comes from belonging to their group or community, not from individual achievements alone. & Society \\ 
& Authority should be respected, and decisions from leaders are not to be questioned & Politics \\ 
& Planning and persistence today are essential to secure the success of future generations. & Morals \\ 
\midrule

\multirow{5}{4cm}{\textbf{Schwartz' Theory of Basic Values} \newline \cite{schwartz2012overview}}
& Life is meant to be full of adventure and new experiences. & Culture \\
& People should be free to think, choose, and act for themselves. & Ethics \\ 
& Success should be measured by what a person achieves through their abilities. & Society \\ 
& A good society protects its people through order, stability, and safety & Politics \\ 
& Respecting traditions and the beliefs of past generations is a moral duty. & Morals \\
\midrule
\multirow{6}{4cm}{\textbf{World Values Survey} \newline \cite{Haerpfer2024-jz}}
& Leisure time is important in life. & Culture \\ 
& I trust people from different nationalities than my own. & Ethics \\ 
& If a woman earns more money than her husband, it's almost certain to cause problems. & Society \\ 
& On the whole, men make better political leaders than women do. & Politics \\ 
& It is a duty towards society to have children. & Morals \\
& Whenever science and religion conflict, religion is always right & Religion \\ 

\bottomrule
\end{tabular}
\vspace{-1mm}
\caption{Additional Values and Beliefs seed statements derived from various cultural frameworks including Hofstede's Cultural Dimensions, Schwartz' Theory of Basic Values, and World Values Survey.}
\label{tab:add_seeds}
\end{table*}

After curation of these seeds, we generate pairs of scenarios to curate a set of value/beliefs scenario candidate pool. We use the following prompt template to generate 3 unique scenarios for each seed: \textit{For the following statement: \{seed statement\}, please generate 3 pairs of role-playing scenarios, where the first scenario illustrates the statement and the second contradicts it. Please limit each scenario to 2-3 sentences and do not repeat scenarios}. Table \ref{tab:example_scenarios} presents example generated scenario pairs, generated from GPT-4o.

\begin{table*}[t]
\centering
\scriptsize
\setlength{\tabcolsep}{6pt} 
\begin{tabular}{>{\raggedright\arraybackslash}p{1cm} 
                >{\raggedright\arraybackslash}p{6cm} 
                >{\raggedright\arraybackslash}p{8cm}}
\toprule
\textbf{Topic} & 
\textbf{Seed Statement} & 
\textbf{Example Scenario Pair} 
\\
\addlinespace
\midrule

\multirow{2}{*}{Culture} 
& \multirow{2}{6cm}{Leisure time is important in life.} 
& \textit{Support}: After finishing work, Maria switches off her laptop and joins her friends for a long walk in the park. She tells them she believes relaxation and fun are just as essential as hard work. \\
& & \textit{No Support}: John works late every evening, skipping outings and hobbies to finish more projects. When asked about taking a break, he insists leisure is a waste of time compared to productivity. \\ 

\midrule

\multirow{2}{*}{Ethics} 
& \multirow{2}{6cm}{I trust people from different nationalities than my own.} 
& \textit{Support}: At an international conference, Aisha gladly shares her research data with a colleague from another country, confident they will use it responsibly. She says collaboration works best when people trust each other across borders. \\ 
& & \textit{No Support}: During a group project, Mark refuses to let a teammate from abroad handle key parts of the work. He mutters that people from other countries can’t be relied on the same way as his own nationals. \\
\midrule 

\multirow{2}{*}{Society} 
& \multirow{2}{6cm}{If a woman earns more money than her husband, it's almost certain to cause problems.} 
& \textit{Support}: At an international conference, Aisha gladly shares her research data with a colleague from another country, confident they will use it responsibly. She says collaboration works best when people trust each other across borders. \\
& & \textit{No Support}: During a group project, Mark refuses to let a teammate from abroad handle key parts of the work. He mutters that people from other countries can’t be relied on the same way as his own nationals. \\ \midrule 

\multirow{2}{*}{Politics} 
& \multirow{2}{6cm}{On the whole, men make better political leaders than women do.} 
& \textit{Support}: During a classroom debate, Alex argues that history shows most successful leaders have been men, so they are naturally better suited for politics. His classmates nod in agreement, saying women are better at supporting roles. \\
& & \textit{No Support}: At a community meeting, Priya points to examples of female leaders who successfully managed crises with empathy and strength. She argues that leadership depends on skill and vision, not gender, and the audience applauds.\\ \midrule 

\multirow{2}{*}{Morals} 
& \multirow{2}{6cm}{It is a duty towards society to have children.} 
& \textit{Support}: At a family gathering, Amina explains to her cousin that she and her husband are eager to start a family soon. She says raising children is their responsibility to continue the community and care for the next generation. \\
& & \textit{No Support}: During a conversation with friends, Daniel mentions he and his partner decided not to have children. He argues that contributing to society can also mean supporting others, mentoring youth, or focusing on community work. \\
\midrule 

\multirow{2}{*}{Religion} 
& \multirow{2}{6cm}{Whenever science and religion conflict, religion is always right.} 
& \textit{Support}: During a family debate about evolution, Fatima says she accepts the religious creation story over scientific explanations. She explains that when science and faith disagree, faith must guide the truth.\\
& & \textit{No Support}: In biology class, Alex learns about the Big Bang and accepts the evidence despite his church teaching otherwise. He tells his classmates that scientific proof carries more weight than religious doctrine in such conflicts.\\

\bottomrule
\end{tabular}
\vspace{-1mm}
\caption{Example scenario pairs for different seed statements across topics. Each seed is expanded into 3 pairs of scenarios, where the first aligns with the statement and the second contradicts it.}
\label{tab:example_scenarios}
\end{table*}

To determine each user's values and beliefs list, we prompt the model with the following prompt: 
\begin{quote}
    \textit{You will be provided with a set of reviews a user has written as well as demographic attributes including age, gender, and country. For the following statement: \{seed statement\}, please select ONE of \{'support', 'no support', 'neutral'\} based on how well the statement reflects the user's beliefs. Here are the list of reviews: \{reviews\} and demographic information: \{demographics\}.}
\end{quote} 
After generation of each user's value and belief system, preference labels are auto-mapped. Each user has up to 450 pairs of values/belief preference pairs. 
 
\personality To generate personality preference pairs, we utilize two existing datasets: \textit{TRAIT} \cite{Lee2024TRait} and \textit{Big5Chat} \cite{li2024big5chat}. In Table \ref{tab:psych_dataset_stats}, we summarize main statistics from these two datasets. We randomly sample 300 total questions (60 questions for each OCEAN trait, 30 from each dataset)\footnote{As \textit{TRAIT} contains 2 answers for each binary label answer, we randomly select one of the answers from each level to generate a preference pair} Table \ref{tab:example_personality_scenarios} contains example questions from each dataset.  

\begin{table*}[t]
\centering
\scriptsize
\setlength{\tabcolsep}{4pt} 
\renewcommand{\arraystretch}{1.1} 
\begin{tabular}{l c c p{3.5cm}}
\toprule
\textbf{Dataset} & \textbf{Total Qs (Big5)} & \textbf{Scenario Format} & \textbf{Answer Format} \\
\midrule
\textbf{TRAIT} \cite{Lee2024TRait} & 5,000 (400 per trait) & Situation-based & Binary (2 answers per low/high) \\
\textbf{Big5Chat} \cite{li2024big5chat} & 100,000 (20K per trait) & Dialogue-based & Binary \\
\bottomrule
\end{tabular}
\caption{Dataset statistics for LLM-based psychometric SJTs.}
\label{tab:psych_dataset_stats}
\end{table*}

\begin{table*}[t]
\centering
\scriptsize
\setlength{\tabcolsep}{4pt} 
\renewcommand{\arraystretch}{1.1} 
\begin{tabularx}{\textwidth}{l l X} 
\toprule
\textbf{Dataset} & \textbf{Trait} & \textbf{Scenario Pair} \\
\midrule
\textbf{TRAIT} & Openness & 
\textit{Q}: I go to the country fair, which is renowned for its vibrant display of local arts and crafts, including paintings, sculptures, and handmade jewelry. The fair is bustling with people of all ages, and the air is filled with the scent of fresh food and the sound of live music. How should I spend my time at the country fair to make the most of this experience? \newline
\textit{High}: Explore each artist's booth to discover unique pieces and engage with the creators about their inspirations. \newline
\textit{Low}: Visit the top-rated food stalls first to enjoy the local flavors that everyone recommends. \\
\textbf{Big5Chat} & Conscientiousness & \textit{Q}: Hey Noelani, I just got the most insane idea for a new stunt! We should totally try it out and see how crazy we can get! I know you'll love it, and I can already imagine you yelling at the top of your lungs! \newline \textit{High}: Whoa, hold up, let's not get ahead of ourselves. I'm all for trying new things, but we need to assess the risks and make sure we're not putting ourselves or others in harm's way. Let's break it down, analyze the potential consequences, and then decide if it's worth pursuing. \newline \textit{Low}: WOOHOO, YEAH, LET'S DO IT! I don't care if it's crazy or not, it sounds like a blast! I'm so down, let's just go for it and see what happens! Who cares about the consequences, we'll figure it out later! \\
\textbf{TRAIT} & Extraversion & \textit{Q}: I am scheduled to lead a team meeting at work where we need to brainstorm and decide on our next project's direction. How should I approach this meeting to make it productive and inclusive? \newline
\textit{High}: Encourage open dialogue by asking each team member to pitch at least one idea, ensuring everyone contributes and feels involved. \newline
\textit{Low}: Prepare a structured agenda with clear objectives and distribute it before the meeting to allow team members to prepare their thoughts.  \\
\textbf{Big5Chat} & Agreeableness & \textit{Q}: Hey, Gillian! Need a hand with something? You looked a bit lost when you got out of your car. Is everything okay? \newline \textit{High}: Thanks so much, Eathan! I appreciate your help. I was just trying to get my bearings. I'm a bit turned around. Could you point me in the direction of the nearest coffee shop? I need a caffeine boost before my meeting. \textit{Low}: What's it to you, Eathan? I don't need your help. I can handle my own problems. And what's with the small talk? You're not getting anything out of this, so just leave me alone. \\
\textbf{TRAIT} & Neuroticism & \textit{Q}: I have been invited to perform at a local community event, but as the day approaches, I become increasingly nervous about performing in front of a crowd. How should I manage my anxiety and prepare for the performance? \newline \textit{High}: You might consider preparing some back-up plans or escape strategies in case the performance feels too overwhelming when you're on stage.\newline 
Low: Focus on the excitement of performing and the joy it brings, rather than the nerves; see this as a chance to grow and have fun.\\
\bottomrule
\end{tabularx}
\caption{Example scenario pairs for personality preference pairs based on LLM-based psychometric SJTs.}
\label{tab:example_personality_scenarios}
\end{table*}

\section{Product Description Personalization}
\label{app:personalization_prompt}

To generate the final personalized book description for each user, we use a templatized prompt based on their demographics:
\begin{quote}
    \textit{You are a \{age\} \{gender\} from \{country\} with \{interests\}. You have \{personality traits\} and \{values\}. Please generate a personalized description of \{book\} with the original description: \{description\}.}
\end{quote}

\section{Experiment Setup}
\label{app:experiment_setup}

\subsection{Finetuning User-Centric LLMs} 
\label{app:our_method}

To finetune \texttt{Llama-3.2-3B-Instruct}, we use DPO and 4-bit quantized LoRA. We preference-tune a single model for all users using the \texttt{TRL} library. We use a final learning rate of: $2 \times 10^{-5}$ and $\beta$ value of: 0.3. We use a batch size of 64 with a gradient accumulation step size of 2 and 3 epochs with early stopping criterion. We train our model on 2 A6000s with DeepSpeed \cite{rajbhandari2020zeromemoryoptimizationstraining} which takes approximately 11 hours. Table \ref{app:baselines} contains our training prompts for preference aligning the model. 

\subsection{Personalized Prompting Strategies} 
\label{app:baselines} 
We employ three kinds of baselines: (1) Prompting-based, (2) SFT, and (3) Preference-Based. For prompting based models, we allow the model to generate 6-8 sentences to keep length similar to original descriptions. Table \ref{tab:app_baselines} provides details on each approach as well as prompt templates. 

For the LaMP-based approach, we first identify the most relevant book descriptions to the target book recommendation. To do this, we first encode all the descriptions of the books a user has reviewed into dense vector representations using a SentenceBERT (\texttt{all-MiniLM-L6-v2}) model. Each description is thus mapped to a high-dimensional embedding capturing its semantic content. Given a target book, we similarly encode its description into the same embedding space and retrieve the top-5 most similar books based on cosine similarity. For efficient similarity search, we leverage FAISS \cite{Douze2024TheFL} to index and query the embeddings. Based on the extracted similar books, we prepend the user's reviews for these books as context in the generation prompt. 

For our SFT approach, we finetune Llama using 4-bit quantized LoRA for 5 epochs with early stopping criterion. We train with a learning rate of $2 \times 10^{-4}$. For our naive preference-based method, we use DPO to align the model to preferences. For each user, we generate a total of $N$ preference pairs (aligned and misaligned book descriptions) based on the books they have reviewed; $N$ refers to the total number of books the user has reviewed. For users with less than 100 reviewed books, we sample $100-N$ books from their top genres and also generate preference pairs for these books. Thus, each user has at a minimum 100 preference pairs. We train the model for a total of 3 epochs with early stopping. We use a learning rate of $2 \times 10^{-5}$. For both training baselines: SFT and DPO methods, we utilize the \texttt{TRL} library, AdamW optimizer, weight decay of 0.01, a cosine learning rate scheduler, and a warmup ratio of 0.05. We also utilize DeepSpeed and 2 A6000's to train the models.

\begin{table*}[t]
\centering
\scriptsize
\setlength{\tabcolsep}{6pt} 
\begin{tabular}{>{\raggedright\arraybackslash}p{2cm} >{\raggedright\arraybackslash}p{13cm} > {\raggedright\arraybackslash}p{2cm}}
\toprule
\textbf{Baseline} & 
\textbf{Prompt Template/Training Example} 
\\
\addlinespace
\midrule

\textit{BaseRewrite} & \textit{Personalization Prompt}: Please generate a more engaging and interesting description for this book: {book} with this description: \{description\}. \\ 

\textit{DemoBased} & \textit{Personalization Prompt}: You are a \{age\} \{gender\} from \{country\}. Please generate a more engaging and interesting description for this book: \{book\} with this description: \{description\}. \\

\textit{UserSummary} & 
\textit{User-Summary Generation Prompt}: 
Please generate a summary of the user based on their historical reviews: [$r_1$, $r_2$, ..., $r_n$]. 

\textit{Personalization Prompt}: Based on this user summary (\textit{user\_summary}), 
please generate a more engaging and interesting description for this book (\textit{book}) 
based on the following description: \textit{description}.
\\

\textit{LaMP} & \textit{Personalization Prompt}: Based on these user reviews: [$r_1, r_2, r_3, r_4, r_5$], please generate a more engaging and interesting description for this book (\textit{book}) 
based on the following description: \textit{description}. \\

\textit{TriAgent} & 
\textit{First Generation Prompt}: Please generate a more engaging and interesting description for this book: \{book\} with this description: \{description\}.\newline
\textit{Edit Instructions Prompt}: Based on this user summary \{user\_summary\} and this personalized book description: \{personalized\_description\}, please generate a set of suggested edits to make the description more engaging and interesting for the user.\newline
\textit{Final Generation Prompt}: Based on this user summary \{user\_summary\} and these suggested edits, please generate a more engaging and interesting description for this book: \{book\} with this description: \{personalized\_description\}. \\
\textit{UserSFT} & \textit{Training Prompt}: You are a \{age\} \{gender\} from \{country\}. You recently read the book: \{book\} with the following description: \{description\} and this was your review: \newline 
\textit{Training Output}: \{review\}\\
\textit{PrefAlign} & \textit{Aligned Description Generation Prompt}: You are a \{age\} \{gender\} from \{country\}. Please generate a more engaging and interesting description for this book: \{book\} with this description: \{description\}. \newline 
\textit{Misaligned Description Generation Prompt}: You are a \{age\} \{gender\} from \{country\}. Please generate a less engaging or interesting description for this book: \{book\} with this description: \{description\}. \newline {DPO Training Prompt}: You are a \{age\} \{gender\} from \{country\}. Which book description is more engaging and interesting? \newline 
\textit{Chosen}: \{Aligned Generated Description\} \newline 
\textit{Rejected}: \{Misaligned Generated Description\} \newline
\textit{Final Description Generation Prompt}: You are a \{age\} \{gender\} from \{country\}. Please generate a more engaging and interesting description for this book: {book} with this description: \{description\}. 
\\  
\textit{\approach} & \textit{User Value Survey}: Given the user's set of values and beliefs, please construct a short paragraph (max. 6 sentences), summarizing the user's values on culture, ethics, society, politics, politics, morals, and religion. \newline 
\textit{DPO Training Prompt}: You are a \{age\} \{gender\} from \{country\}. You have the following values: \{user\_value\_summary\} and personality traits: \{traits\}. \{preference\_prompt\} \newline 
\textit{Chosen}: \{preference\_prompt\_chosen\} \newline 
\textit{Rejected}: \{preference\_prompt\_rejected\} \newline 
\textit{Final Description Generation Prompt}: You are a \{age\} \{gender\} from \{country\}. You have the following values: \{user\_value\_summary\} and personality traits: \{traits\}. Please generate a more engaging and interesting description for this book: {book} with this description: \{description\}. \\
\bottomrule
\end{tabular}
\vspace{-1mm}
\caption{All baselines with detailed prompt templates and training examples.}
\label{tab:app_baselines}
\end{table*}

\subsection{Auto-Evaluation Metrics} 
\label{app:metrics}

For evaluation, we utilize GPT-4o as an LLM-judge. To simulate a user, we generate a persona prompt that includes their demographics (age, gender, and country) as well as a user summary, which is generated by GPT-4o from all of the user's historical reviews. For each user, we generate a total of five personalized book descriptions, resulting in 2,000 generations per method. The LLM-judge is then provided with all personalized generations and asked to rank them according to engagement and interest using the following prompt:

\begin{quote}
    \textit{You are a \{age\} \{gender\} from \{country\}. \{user\_summary\}. Please carefully read each of the following book descriptions and provide a ranking of how engaging and interesting you find each description: [description\_0\footnote{This refers to the original book description}, description\_1, description\_2, description\_3, description\_4, description\_5, description\_6, description\_7].}
\end{quote}

We then calculate the \textit{WinRate} for each method by computing the percentage of times the model achieves first place compared to all other methods across all users and generations. We also compute \textit{Preference Gain} by calculating the percentage of times the model's output is preferred over the original description. For measuring \textit{Interestingness}, we prompt the model one description at a time using the following Likert scale (1-5): 1 = not very interesting/engaging, 5 = very interesting/engaging. 

\begin{quote}
    \textit{You are a \{age\} \{gender\} from \{country\}. \{user\_summary\}. Please carefully read the following book description and rate how engaging and interesting you find the description from 1 to 5: \{description\}.}
\end{quote}

\subsection{User Study Design}
\label{app:user_study}

\begin{figure*}[!t] \centering \includegraphics[width=0.6\textwidth]{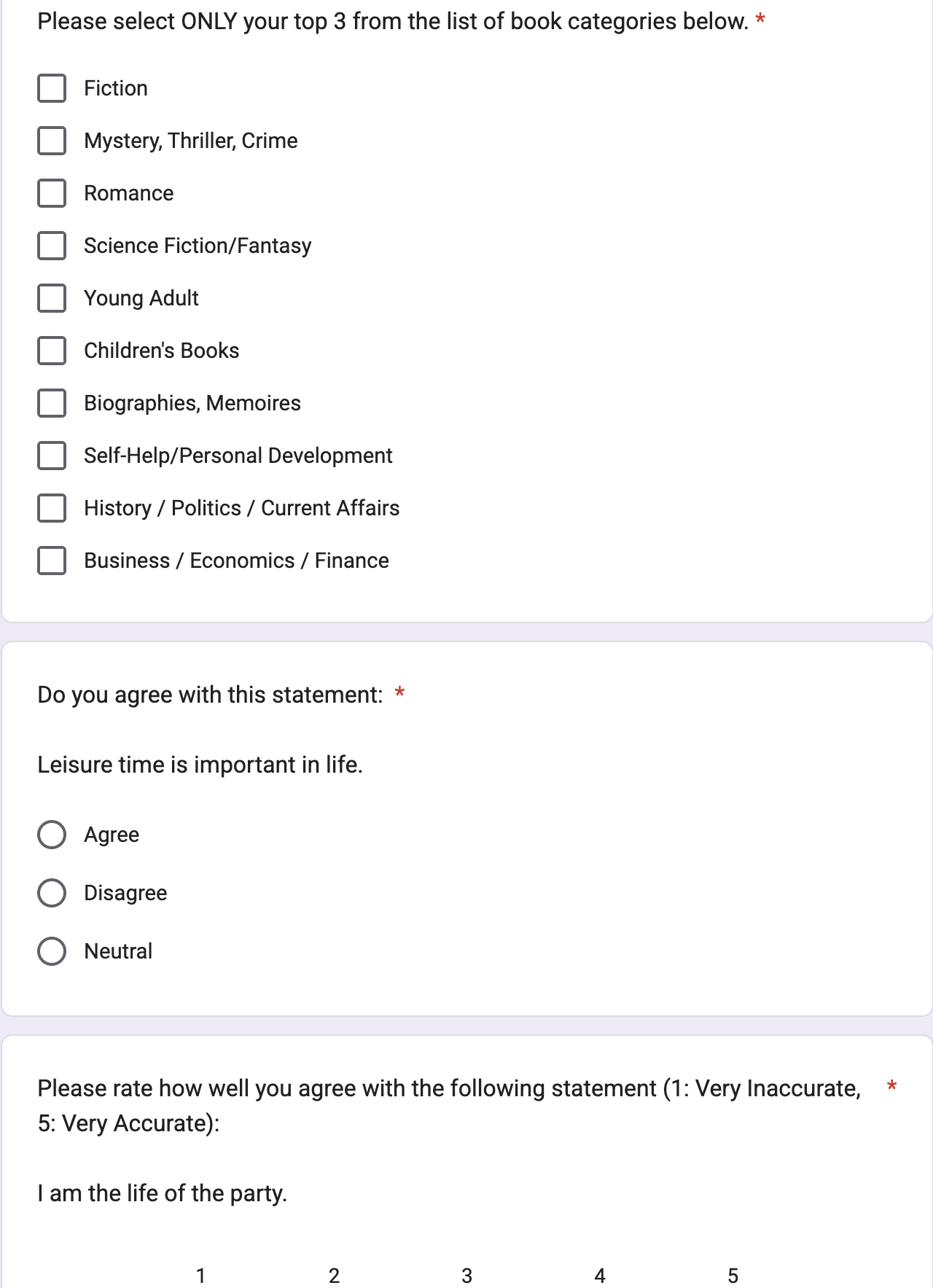} \caption{Example questions from first stage in user study: data collection. Along with standard demographics, we extract user interests i.e. ranking top 3 genres, value systems, and personality.} 
\label{fig:study_p1} \end{figure*}

We conduct a multi-part user study on Prolific to evaluate personalized generation across diverse cultural backgrounds. We recruit 30 participants each from the United States, Brazil, Japan, and India. In the first phase (see Figure~\ref{fig:study_p1}), we extract user attributes including gender and age group (as defined in \S\ref{sec:user_profiles}). To capture users' literary preferences, participants select their top three book genres from a list of nine categories\footnote{We adopt the hierarchical genre structure shown in Figure~\ref{fig:genre_experiments}.}. 

We further measure users’ personality traits using the Mini-IPIP questionnaire~\cite{donnellan2006mini}, a validated 20-item Likert-scale instrument for assessing the Big Five (OCEAN) traits. To estimate users’ values and beliefs, we adapt our 150-item seed statement bank, selecting a concise subset of 10 statements, across various categories: culture, ethics, society, politics, morals, and religion, to minimize participant fatigue. We then prompt GPT-4o to infer users’ likely responses to the remaining statements based on their annotated subset:
\begin{quote}
\textit{Given a user's value system: \{annotated seed statements\}, would the user agree, disagree, or be neutral to this statement: \{additional seed statement\}?}
\end{quote}

\begin{figure*}[!t] \centering \includegraphics[width=0.6\textwidth]{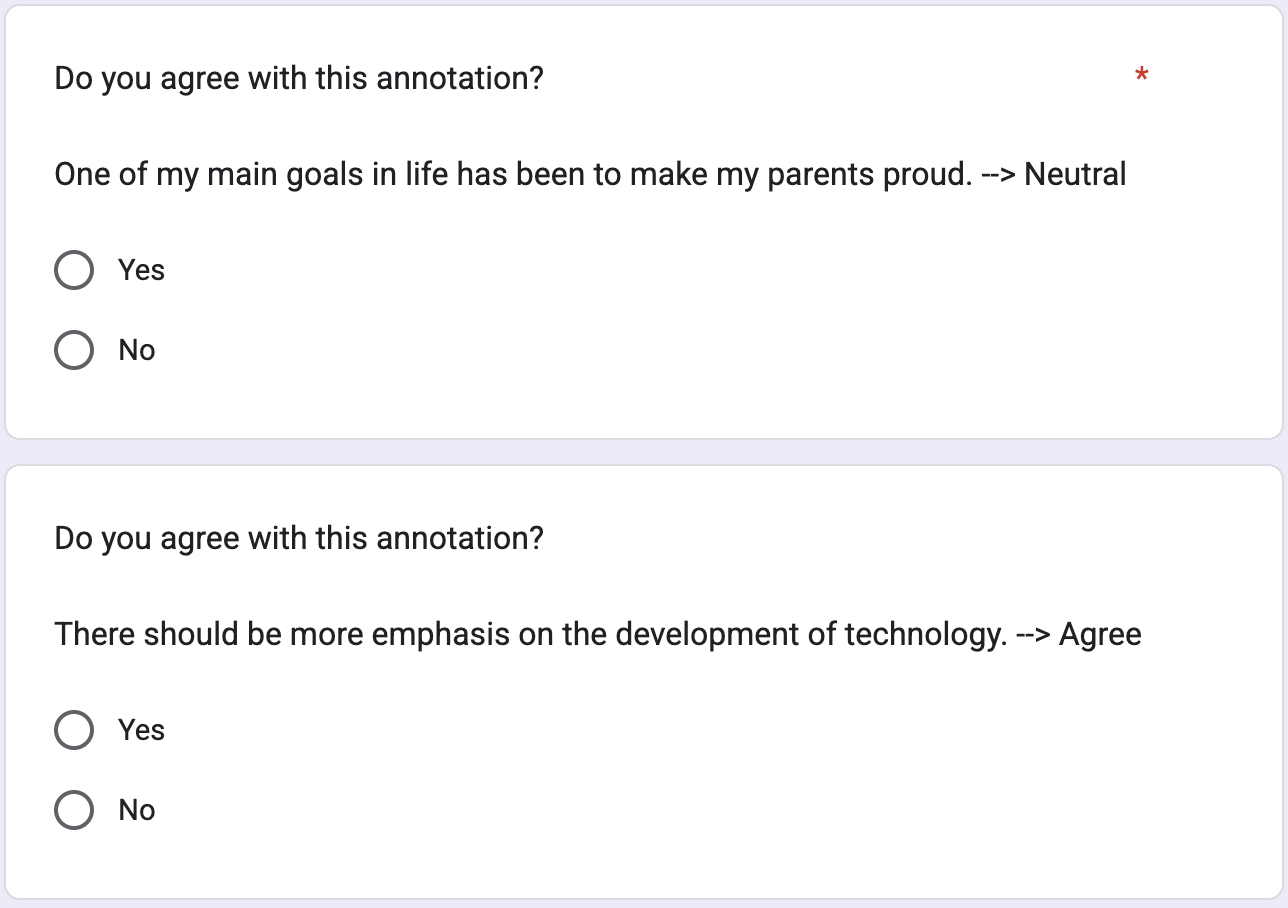} \caption{Example questions from second stage in user study: user values verification. Users are given additional values seed statements and asked to verify whether they agree with GPT-4o annotations.} 
\label{fig:study_p2} \end{figure*}

To validate GPT-4o’s value inference accuracy, in Step~2 of our study (Figure~\ref{fig:study_p2}), each participant is presented with five additional seed statements and asked to indicate if they agree with the GPT-4o annotation label. Across all four nations, GPT-4o achieves an average accuracy of 84\%, indicating that the model can effectively infer user value systems from limited supervision (10 examples per user).

\begin{figure*}[!t] \centering \includegraphics[width=0.6\textwidth]{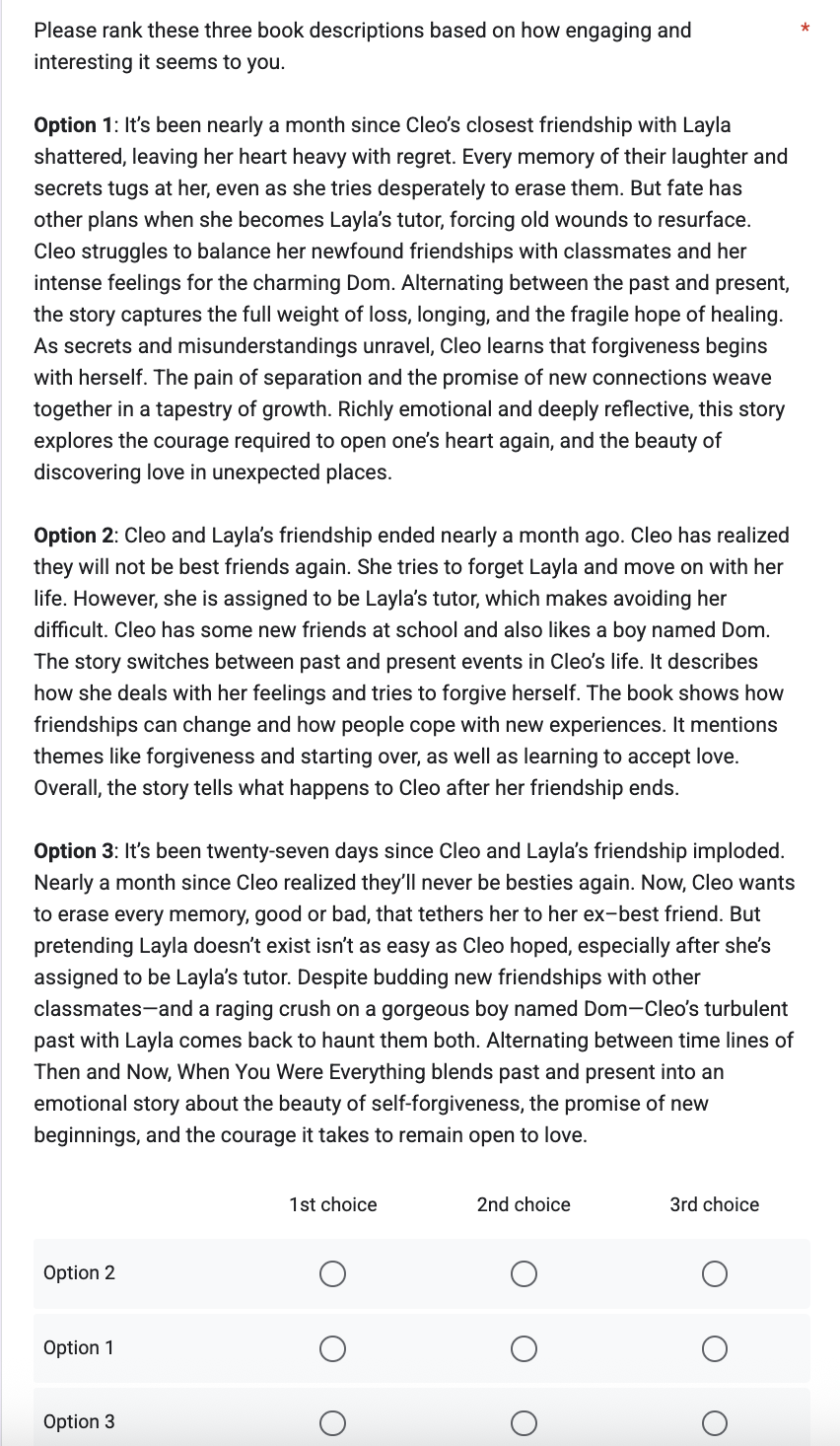} \caption{Example questions from second stage in user study: user values verification. Users are given additional values seed statements and asked to verify whether they agree with GPT-4o annotations.} 
\label{fig:study_p3} \end{figure*}

In the final phase of our study, each participant is shown ten triplets of book descriptions: the original, one generated by \approach, and one by a strong baseline model (\textit{TriAgent}), which does not rely on additional user or synthetic data. Participants are asked to (1) rank the descriptions by engagement and interest, (2) rate each on a 1–5 Likert scale for interestingness, and (3) note portions they find particularly appealing (Figure~\ref{fig:study_p3}). Table~\ref{tab:user_examples} illustrates an example annotation. Overall, users consistently report that \approach\ outputs align more closely with their cultural norms, personal experiences, and value systems, underscoring the importance of integrating culture and values into personalization. 

\begin{table*}[t]
\centering
\scriptsize
\setlength{\tabcolsep}{4pt} 
\renewcommand{\arraystretch}{1.2} 
\begin{tabular}{p{4cm} p{10.5cm}}
\toprule
\textbf{User Details} & \textbf{User Highlights and Rationales} \\
\midrule

\begin{minipage}[t]{4cm}
\raggedright
\textit{Country}: Brazil \\ 
\textit{Age}: Middle-Aged \\ 
\textit{Gender}: Female \\ 
\textit{Value}: strong supporter of women's rights and abortion
\end{minipage} 
&
\begin{minipage}[t]{10.5cm}
\raggedright
\textit{Highlight}: When Nnu discovers she must make a life-changing choice, she confronts the pressure of family expectations and societal judgment, ultimately finding the courage to take control of her own body and destiny, standing firm in the belief that her choices belong to her alone. \\
\textit{Rationale}: Feminism is something that I have been passionate about for many years now. I felt like it captured exactly what it’s like to make your own choices, even when everyone around you has opinions.
\end{minipage} 
\\
\midrule 

\begin{minipage}[t]{4cm}
\raggedright
\textit{Country}: USA \\ 
\textit{Age}: Senior \\ 
\textit{Gender}: Female \\ 
\textit{Value}: life should be a mixture of fun and work
\end{minipage} 
&
\begin{minipage}[t]{10.5cm}
\raggedright
\textit{Highlight}: She discovered that the secret to happiness wasn’t escaping responsibility, but weaving laughter and discovery into every day’s obligations. \\
\textit{Rationale}: I think the same way as the main character here, so it probably resonated most with me.
\end{minipage} 
\\

\begin{minipage}[t]{4cm}
\raggedright
\textit{Country}: India \\ 
\textit{Age}: Middle-Aged \\ 
\textit{Gender}: Male \\ 
\textit{Personality}: high Neuroticism and low Extraversion 
\end{minipage} 
&
\begin{minipage}[t]{10.5cm}
\raggedright
\textit{Highlight}: Alone in his dimly lit apartment, Ronald felt the walls closing in as the thought of his feelings became louder and he began pondering how he would escape from… \\
\textit{Rationale}: The way this was written somehow resonated well with me, maybe because I think I might respond in a similar manner in a given situation.
\end{minipage} 
\\
\midrule 

\begin{minipage}[t]{4cm}
\raggedright
\textit{Country}: Japan \\ 
\textit{Age}: Senior \\ 
\textit{Gender}: Male \\ 
\textit{Value}: supporter of culture and religion \\
\textit{Personality}: high Openness 
\end{minipage} 
&
\begin{minipage}[t]{10.5cm}
\raggedright
\textit{Highlight}: Learn more about various Aztec festivals, including Toxcatl, one of the largest festivals devoted to Tezcatlipoca, the god of the… \\
\textit{Rationale}: This was particularly interesting to me which none of the other descriptions highlighted that well because I’m very interested in different cultures and this one highlighted an intriguing old celebration. 
\end{minipage} 
\\

\bottomrule
\end{tabular}
\caption{Example highlights users selected from \approach\ generations and reasons/rationales for why the description was more engaging and interesting to them. Many users highlight notions from their cultural, intrinsic values/beliefs, and personality/personal experiences.}
\label{tab:user_examples}
\end{table*}

\section{Additional Results}

In this section, we present personalization metrics (Top-1 WinRates, Preference Gains, Interestingness Scores) split by the Amazon readers' country. We find that split across various countries, \approach generations can yield substantial yields especially in Non-Western settings (see Tables \ref{tab:usa_results}, \ref{tab:brazil_results}, \ref{tab:india_results}, and \ref{tab:japan_results}).
\begin{table*}
\centering
\scriptsize
\begin{tabular}{
    >{\centering\arraybackslash}m{4.5cm}  
    >{\centering\arraybackslash}m{2.25cm}  
    >{\centering\arraybackslash}m{2.25cm}  
    >{\centering\arraybackslash}m{3cm}  
}
\toprule
\textbf{Personalization Method} & \textbf{Top-1 WinRate (\%)} & \textbf{Preference Gain (\%)} & \textbf{Interestingness Score} \\
\midrule
\textit{Original} & 0.75 & - & 3.65 \\
\textit{BaseRewrite} & 1.75 & 68.0 & 3.72 \\
\textit{DemoBased} & 3.5 & 76.25 & 3.60 \\
\textit{UserSummary} & 9.75 & 74.0 & 3.92 \\
\textit{LaMP} & 13.25 & 79.75 & \colorbox{lightgreen}{4.00}  \\
\textit{TriAgent} & 7.5 & 78.25 & 3.88 \\
\textit{UserSFT} & 21.25 & 77.75 & 3.85  \\
\textit{PrefAlign} & 21.0 & 79.5 & 3.95 \\
\approach (Ours) & \colorbox{lightgreen}{26.25} & \colorbox{lightgreen}{80.75} & 3.97 \\
\bottomrule
\end{tabular}
\vspace{-1mm}
\caption{Automatic personalization metrics (GPT-4o evaluated) for American Amazon readers.}
\label{tab:usa_results}
\end{table*}

\begin{table*}
\centering
\scriptsize
\begin{tabular}{
    >{\centering\arraybackslash}m{4.5cm}  
    >{\centering\arraybackslash}m{2.25cm}  
    >{\centering\arraybackslash}m{2.25cm}  
    >{\centering\arraybackslash}m{3cm}  
}
\toprule
\textbf{Personalization Method} & \textbf{Top-1 WinRate (\%)} & \textbf{Preference Gain (\%)} & \textbf{Interestingness Score} \\
\midrule
\textit{Original} & 0.5 & - & 3.75 \\
\textit{BaseRewrite} & 1.0 & 68.0 & 3.80 \\
\textit{DemoBased} & 3.75 & 68.5 & 3.72 \\
\textit{UserSummary} & 10.25 & 73.0 & 3.99 \\
\textit{LaMP} & 5.5 & 78.75 & \colorbox{lightgreen}{4.05}  \\
\textit{TriAgent} & 7.75 & 76.25 & 3.94 \\
\textit{UserSFT} & 20.5 & 75.5 & 3.90 \\
\textit{PrefAlign} & 24.5 & \colorbox{lightgreen}{81.25} & 4.02  \\
\approach (Ours) & \colorbox{lightgreen}{27.25} & 78.5 & 4.03  \\
\bottomrule
\end{tabular}
\vspace{-1mm}
\caption{Automatic personalization metrics (GPT-4o evaluated) for Brazil Amazon readers}
\label{tab:brazil_results}
\end{table*}

\begin{table*}
\centering
\scriptsize
\begin{tabular}{
    >{\centering\arraybackslash}m{4.5cm}  
    >{\centering\arraybackslash}m{2.25cm}  
    >{\centering\arraybackslash}m{2.25cm}  
    >{\centering\arraybackslash}m{3cm}  
}
\toprule
\textbf{Personalization Method} & \textbf{Top-1 WinRate (\%)} & \textbf{Preference Gain (\%)} & \textbf{Interestingness Score}  \\
\midrule
\textit{Original} & 0.75 & - & 3.80 \\
\textit{BaseRewrite} & 0.75 & 60.25 & 3.85 \\
\textit{DemoBased} & 4 & 70.75 & 3.72 \\
\textit{UserSummary} & 7.0 & 68.75 & 4.03\\
\textit{LaMP} & 12.5 & 70.5 & 4.08  \\
\textit{TriAgent} & 6.5 & 69.25 & 3.97  \\
\textit{UserSFT} & 17.75 & 71.25 & 3.92  \\
\textit{PrefAlign} & 21.25 & 74.5 & 4.04 \\
\approach (Ours) & \colorbox{lightgreen}{28.0} & \colorbox{lightgreen}{86.25} & \colorbox{lightgreen}{4.10} \\
\bottomrule
\end{tabular}
\vspace{-1mm}
\caption{Automatic personalization metrics (GPT-4o evaluated) for Japanese Amazon readers.}
\label{tab:japan_results}
\end{table*}

\begin{table*}
\centering
\scriptsize
\begin{tabular}{
    >{\centering\arraybackslash}m{4.5cm}  
    >{\centering\arraybackslash}m{2.25cm}  
    >{\centering\arraybackslash}m{2.25cm}  
    >{\centering\arraybackslash}m{3cm}  
}
\toprule
\textbf{Personalization Method} & \textbf{Top-1 WinRate (\%)} & \textbf{Preference Gain (\%)} & \textbf{Interestingness Score} \\
\midrule
\textit{Original} & 1.0 & - & 3.76 \\
\textit{BaseRewrite} & 1.5 & 60.75 & 3.87 \\
\textit{DemoBased} & 1.75 & 59.25 & 3.74 \\
\textit{UserSummary} & 10.0 & 72.5 & 4.06 \\
\textit{LaMP} & 8.75 & 70.25 & \colorbox{lightgreen}{4.11} \\
\textit{TriAgent} & 6.25 & 72.25 & 3.96  \\
\textit{UserSFT} & 18.5 & 74.5 & 3.97  \\
\textit{PrefAlign} & 19.5 & 74.25 & 4.08  \\
\approach (Ours) & \colorbox{lightgreen}{28.25} & \colorbox{lightgreen}{83.75} & 4.10  \\
\bottomrule
\end{tabular}
\vspace{-1mm}
\caption{Automatic personalization metrics (GPT-4o evaluated) for Indian Amazon readers.}
\label{tab:india_results}
\end{table*}

\end{document}